\documentclass[fleqn,twoside]{article}
\usepackage{espcrc1}

\newcommand{\AmS}{{\protect\the\textfont2
  A\kern-.1667em\lower.5ex\hbox{M}\kern-.125emS}}

\title{Combinatorial framework for planning in geological exploration}

\author{Mark Sh. Levin}

\begin{document}

\maketitle

\begin{abstract}
The paper describes combinatorial framework
 for planning of geological exploration for oil-gas fields.
 The suggested scheme of the  geological exploration
 involves the following stages:
 (1) building of special 4-layer tree-like model
  (layer of geological exploration):
  productive layer,
  group of productive layers,
  oil-gas field,
  oil-gas region (or group of the fields);
 (2) generations of local design (exploration) alternatives
 for each low-layer geological objects:
  conservation, additional search, independent utilization,
  joint utilization;
 (3) multicriteria (i.e., multi-attribute)
  assessment of the design (exploration) alternatives and their
  interrelation (compatibility) and mapping if the obtained
  vector estimates into integrated ordinal scale;
 (4) hierarchical design ('bottom-up') of composite exploration
 plans for each oil-gas field;
 (5) integration of the plans into region plans;
  and
 (6) aggregation of the region plans into a general exploration
  plan.
 Stages 2, 3, 4, and 5 are based on
 hierarchical multicriteria morphological design (HMMD) method
  (assessment of ranking of alternatives, selection and composition
  of alternatives into composite alternatives).
  The composition problem is based on morphological clique model.
 Aggregation of the obtained modular alternatives (stage 6)
 is based on detection of a alternatives 'kernel'
 and its extension by addition of elements (multiple choice
 model).
 In addition,
 the usage of multiset estimates for alternatives is
 described as well.
 The alternative estimates are based on expert judgment.
 The suggested combinatorial planning methodology
 is illustrated by numerical examples for geological
 exploration of Yamal peninsula.

~~


\vspace{1pc}
\end{abstract}

\maketitle

\tableofcontents

\newcounter{cms}
\setlength{\unitlength}{1mm}

\newpage
\section{Introduction}

 In recent decades significance of
 mineral resources and their geological exploring
 has been increased.
 This paper is focusing on combinatorial planning
 of geological exploration for oil and gas fields.
 The suggested framework is the following (Fig. 1):

  {\it Stage 1.}
 An analysis of the initial applied problem
 and a preliminary its structuring (for example,
  partitioning the problem into parts, generation of basic
  requirements/criteria).

 {\it Stage 2.}
 Designing a special four-layer tree-like model
 (as a multi-layer model of geological objects):
 (i) productive stratum (reservoir),
 (ii) group of productive stratums (reservoirs)),
 (iii) oil and gas field, and
 (iv) group of oil and gas fields (region).

  {\it Stage 3.}
   Generation for each bottom-layer geological object
 a set of local design alternatives  (for geological exploration) DAs:
 (a) conservation,
 (b) appraisal work, 
 (c) independent production time, 
 (d) joint independent production time, 

 {\it Stage 4.} Multicriteria assessment of DAs and their
  interconnection (compatibility IC)
 and mapping the obtained vector estimates into an ordinal scale.

 {\it Stage 5.} Hierarchical (bottom-up) composition of
 composite exploration plans for each field.

  {\it Stage 6.}
 Integration of the obtained plans for the fields
 into region plans.

 {\it Stage 7.} Aggregation of the obtained
 region plans (solutions).

\begin{center}
\begin{picture}(72,93.5)
\put(04,00){\makebox(0,0)[bl]{Fig. 1.
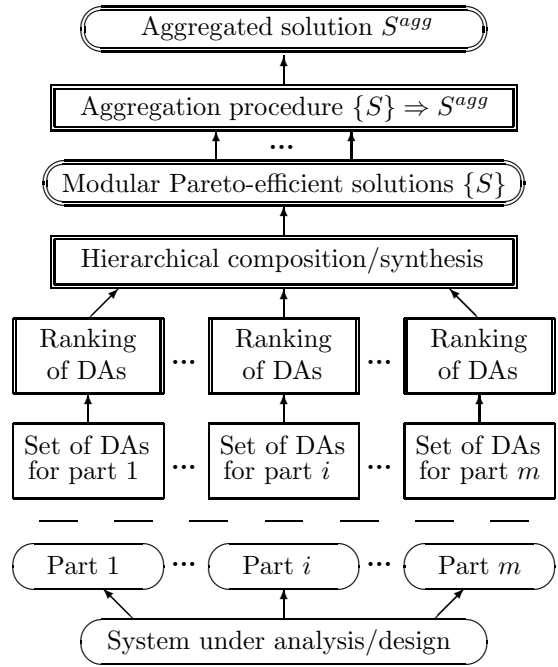
 General combinatorial framework}}

\put(36,89.5){\oval(54,6)} \put(36,89.5){\oval(53,5)}
\put(17,88){\makebox(0,0)[bl]{Aggregated solution \(S^{agg}\)}}

\put(05,76){\line(1,0){62}} \put(05,82){\line(1,0){62}}
\put(05,76){\line(0,1){06}} \put(67,76){\line(0,1){06}}

\put(05.5,76.5){\line(1,0){61}} \put(05.5,81.5){\line(1,0){61}}
\put(05.5,76.5){\line(0,1){05}} \put(66.5,76.5){\line(0,1){05}}

\put(08.9,77.2){\makebox(0,0)[bl]{Aggregation procedure
  \(\{S\} \Rightarrow S^{agg}\) }}

\put(36,82){\vector(0,1){04}}

\put(36,69){\oval(64,6)} \put(36,69){\oval(63,5)}

\put(06.5,67){\makebox(0,0)[bl]{Modular Pareto-efficient
 solutions  \(\{S\}\)}}

\put(27,72){\vector(0,1){04}}

\put(34,73.5){\makebox(0,0)[bl]{{\bf ...}}}

\put(45,72){\vector(0,1){04}}

\put(05,55){\line(1,0){62}} \put(05,62){\line(1,0){62}}
\put(05,55){\line(0,1){07}} \put(67,55){\line(0,1){07}}

\put(05.5,55.5){\line(1,0){61}} \put(05.5,61.5){\line(1,0){61}}
\put(05.5,55.5){\line(0,1){06}} \put(66.5,55.5){\line(0,1){06}}

\put(09,57){\makebox(0,0)[bl]{Hierarchical composition/synthesis}}

\put(36,62){\vector(0,1){04}}

\put(00,41){\line(1,0){20}} \put(00,51){\line(1,0){20}}
\put(00,41){\line(0,1){10}} \put(20,41){\line(0,1){10}}

\put(0.4,41){\line(0,1){10}} \put(19.6,41){\line(0,1){10}}

\put(03.5,46.5){\makebox(0,0)[bl]{Ranking}}
\put(04.5,43){\makebox(0,0)[bl]{of DAs }}

\put(10,51){\vector(1,1){4}}

\put(21,45){\makebox(0,0)[bl]{{\bf ...}}}
\put(47,45){\makebox(0,0)[bl]{{\bf ...}}}

\put(26,41){\line(1,0){20}} \put(26,51){\line(1,0){20}}
\put(26,41){\line(0,1){10}} \put(46,41){\line(0,1){10}}

\put(26.4,41){\line(0,1){10}} \put(45.6,41){\line(0,1){10}}

\put(29.5,46.5){\makebox(0,0)[bl]{Ranking}}
\put(30.5,43){\makebox(0,0)[bl]{of DAs}}

\put(36,51){\vector(0,1){4}}

\put(52,41){\line(1,0){20}} \put(52,51){\line(1,0){20}}
\put(52,41){\line(0,1){10}} \put(72,41){\line(0,1){10}}

\put(52.4,41){\line(0,1){10}} \put(71.6,41){\line(0,1){10}}

\put(55.5,46.5){\makebox(0,0)[bl]{Ranking}}
\put(56.5,43){\makebox(0,0)[bl]{of DAs}}

\put(62,51){\vector(-1,1){4}}

\put(00,27){\line(1,0){20}} \put(00,37){\line(1,0){20}}
\put(00,27){\line(0,1){10}} \put(20,27){\line(0,1){10}}

\put(01.5,33){\makebox(0,0)[bl]{Set of DAs}}
\put(02,29){\makebox(0,0)[bl]{for part \(1\)}}

\put(10,37){\vector(0,1){4}}

\put(21,31){\makebox(0,0)[bl]{{\bf ...}}}
\put(47,31){\makebox(0,0)[bl]{{\bf ...}}}

\put(26,27){\line(1,0){20}} \put(26,37){\line(1,0){20}}
\put(26,27){\line(0,1){10}} \put(46,27){\line(0,1){10}}

\put(27.5,33){\makebox(0,0)[bl]{Set of DAs}}
\put(28,29){\makebox(0,0)[bl]{for part \(i\) }}

\put(36,37){\vector(0,1){4}}

\put(52,27){\line(1,0){20}} \put(52,37){\line(1,0){20}}
\put(52,27){\line(0,1){10}} \put(72,27){\line(0,1){10}}

\put(53.5,33){\makebox(0,0)[bl]{Set of DAs}}
\put(54,29){\makebox(0,0)[bl]{for part \(m\)}}

\put(62,37){\vector(0,1){4}}

\put(03.5,24){\line(1,0){5}} \put(13.5,24){\line(1,0){5}}
\put(23.5,24){\line(1,0){5}} \put(33.5,24){\line(1,0){5}}
\put(43.5,24){\line(1,0){5}} \put(53.5,24){\line(1,0){5}}
\put(63.5,24){\line(1,0){5}}

\put(10,18){\oval(20,6)}

\put(04.5,17){\makebox(0,0)[bl]{Part \(1\)}}

\put(36,18){\oval(20,6)}

\put(30.5,17){\makebox(0,0)[bl]{Part \(i\)}}

\put(62,18){\oval(20,6)}

\put(56.5,17){\makebox(0,0)[bl]{Part \(m\)}}

\put(21,18){\makebox(0,0)[bl]{{\bf ...}}}
\put(47,18){\makebox(0,0)[bl]{{\bf ...}}}

\put(16,11){\vector(-1,1){4}} \put(36,11){\vector(0,1){4}}
\put(56,11){\vector(1,1){4}}
\put(36,08){\oval(54,6)}

\put(12.5,6.1){\makebox(0,0)[bl]{System under analysis/design}}

\end{picture}
\end{center}

 Stages 2, 3, 4, 5, 6 are based on hierarchical morphological
 multicriteria design (HMMD) method
 (ranking of DAs, selection and composition of DAs)
 \cite{lev96,lev98,lev06b,lev15}.
 Stage 7 consists in aggregation of the obtained modular solutions
 (detection of a 'kernel' of the obtained solutions and its extension
 by some additional solution elements)
  \cite{lev11agg,lev15}.
 An  example of combinatorial solution based on
 multiset estimates of DAs is described as well.
 All stages above are based on expert judgment.
 Note, combinatorial approach for selection of optimal
 geological actions based on knapsack-like model was described in
 \cite{koltun05}.

 The suggested combinatorial framework is illustrated
 by a numerical example for Yamal peninsula.
 The preliminary compressed material was published
 in \cite{levpor97}.
 Mainly, initial  information is based on handbook
   \cite{maximov79} and expert judgment of Vladimir I. Poroskun.
 A preliminary Russian version of the material is contained in
 \cite{lev13b}.

\section{Description of framework parts}

 Here brief descriptions of the suggested framework parts
  are presented (as a simplified introduction for readers).

\subsection{Hierarchical morphological design}

 Hierarchical morphological multicriteria design (HMMD) method
 has been described in several publications
 \cite{lev96,lev98,lev06b,lev15}.

\subsubsection{Hierarchical system model}

 In HMMD a special hierarchical (tree-like)
 model of the analyzed system
 'morphological tree' is used:
 \cite{lev96,lev98,lev06b,lev15}:
 (i) tree-like system model,
 (ii) set of leaf node as bottom-layer components (parts)
 of the systems,
 (iii) set of design alternatives (DAs)
 for each bottom system component,
 (iv) ranking of DAs for each each bottom system component
 (to obtain an ordinal estimate/priority for each DA),
 (v) ordinal estimates of compatibility between
  DAs of neighbor system components.
%

\subsubsection{Ordinal estimates of system components}

 Here the basic version of HMMD (ordinal estimated of DAs)
 is briefly described.
 The following is assumed:
 (1) system quality is considered a two-component estimate:
   quality of components and quality of their compatibility;
 (2) monotone criteria for the system and its parts are
 considered;
 (3) an ordinal scale is used for quality of system components
    (i.e., local solutions); and
 (4) an ordinal scale is used for quality
 of system component compatibility.
 The following designations are used:
  (a) priorities (ordinal estimates) for design alternatives (DAs):
  \(\iota=1,...,l\),   \(1\) corresponds the the best quality level;
  (b) ordinal compatibility for pair of DAs:
 \(w=0,...,\nu\),
 \(0\) corresponds to impossible (the worst) quality level.

 Let \(S\) be a composite solution (DA) consisting of \(m\) parts.
 The synthesis problem is based on morphological clique model:

~~

 Find composite system
 ~~\(S=S(1)\star ...\star S(i)\star ...\star S(m)\),~
 consisting of parts/components (i.e., local DAs)
 (one representative \(S(i)\) for
 \(i\)-th system component \(i=1,...,m\))
 with non-zero compatibility estimates between the selected
 pair of DAs.

~~

 The poset of the system quality for composite solution \(S\)
  is based on vector:
 ~~\(N(S)=(w(S);e(S))\),~ where \(w(S)\) corresponds
 to minimum of compatibility estimates for DAs pair in \(S\),
 \(e(S)=(\eta_{1},...,\eta_{\iota},...,\eta_{l})\),
 where \(\eta_{\iota}\) corresponds to the number of local DAs at quality level
  \(\iota\) in \(S\).
  Two-criteria optimization model is:
%
\[ \max ~ e(S), ~~~ \max ~ w(S), ~~~ s.t. ~~ w(S) \geq 1.\]
 As a result, non-dominated by
  \(N(S)\) (Pareto-efficient) composite solutions are search for.
 The model belongs to class of NP-hard problems.
 The evident solving scheme involves two stages:

~~

 {\it Stage 1.} Building of all admissible solutions
  (composite DAs);

 {\it Stage 2.} Selection of Pareto-efficient solutions.

~~

 Two algorithms can be used for the problem
 \cite{lev96,lev98,lev06b}:
 (1) directed enumeration of solutions
 (start solution(s) corresponds to the best quality estimate(s)),
 search
 (2) dynamic programming based method
 (series construction of admissible solutions for system parts).
 Note, in the case of a small degree of the system tree-based model
 (for example, \([3...7]\))
 algorithmic complexity of the first algorithm is sufficiently
 small.
%
 An illustrative example of tree-component system
  ~\(S=H\star B \star V\) is depicted in Fig. 2
 (priorities of DAs are shown in parentheses).
 The following solutions can be considered:

 (a) \(S_{1} = H_{1}\star B_{1}\star V_{2}\),
  ~\(N(S_{1})=(3;1,1,1)\);

 (b) \(S_{2} = H_{2}\star B_{3}\star V_{2}\),
 ~\(N(S_{2})=(2;2,1,0)\);

 (c) \(S_{3} = H_{3}\star B_{2}\star V_{1}\),
  ~\(N(S_{3})=(1;3,0,0)\).

\begin{center}
\begin{picture}(50,48)
\put(07.5,00){\makebox(0,0)[bl] {Fig. 2.
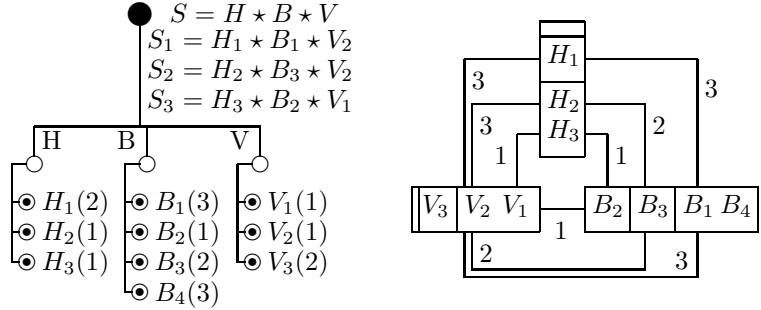
 Illustrative example of combinatorial synthesis}}

\put(4,09){\makebox(0,8)[bl]{\(H_{3}(1)\)}}
\put(4,13){\makebox(0,8)[bl]{\(H_{2}(1)\)}}
\put(4,17){\makebox(0,8)[bl]{\(H_{1}(2)\)}}

\put(19,05){\makebox(0,8)[bl]{\(B_{4}(3)\)}}
\put(19,09){\makebox(0,8)[bl]{\(B_{3}(2)\)}}
\put(19,13){\makebox(0,8)[bl]{\(B_{2}(1)\)}}
\put(19,17){\makebox(0,8)[bl]{\(B_{1}(3)\)}}

\put(34,09){\makebox(0,8)[bl]{\(V_{3}(2)\)}}
\put(34,13){\makebox(0,8)[bl]{\(V_{2}(1)\)}}
\put(34,17){\makebox(0,8)[bl]{\(V_{1}(1)\)}}
\put(3,24){\circle{2}} \put(18,24){\circle{2}}
\put(33,24){\circle{2}}

\put(0,24){\line(1,0){02}} \put(15,24){\line(1,0){02}}
\put(30,24){\line(1,0){02}}

\put(0,24){\line(0,-1){13}} \put(15,24){\line(0,-1){17}}
\put(30,24){\line(0,-1){13}}

\put(30,19){\line(1,0){01}} \put(30,15){\line(1,0){01}}
\put(30,11){\line(1,0){01}}

\put(32,19){\circle{2}} \put(32,19){\circle*{1}}
\put(32,15){\circle{2}} \put(32,15){\circle*{1}}
\put(32,11){\circle{2}} \put(32,11){\circle*{1}}

\put(15,07){\line(1,0){01}} \put(15,11){\line(1,0){01}}
\put(15,15){\line(1,0){01}} \put(15,19){\line(1,0){01}}

\put(17,15){\circle{2}} \put(17,15){\circle*{1}}
\put(17,11){\circle{2}} \put(17,19){\circle{2}}
\put(17,11){\circle*{1}} \put(17,19){\circle*{1}}
\put(17,07){\circle{2}} \put(17,07){\circle*{1}}

\put(0,11){\line(1,0){01}} \put(0,15){\line(1,0){01}}
\put(0,19){\line(1,0){01}}

\put(2,15){\circle{2}} \put(2,19){\circle{2}}
\put(2,15){\circle*{1}} \put(2,19){\circle*{1}}
\put(2,11){\circle{2}} \put(2,11){\circle*{1}}
\put(3,29){\line(0,-1){04}} \put(18,29){\line(0,-1){04}}
\put(33,29){\line(0,-1){04}}

\put(3,29){\line(1,0){30}} \put(17,29){\line(0,1){15}}

\put(17,44){\circle*{3}}

\put(4,26){\makebox(0,8)[bl]{H}} \put(14,26){\makebox(0,8)[bl]{B}}
\put(29,26){\makebox(0,8)[bl]{V}}

\put(21,43){\makebox(0,8)[bl]{\(S = H\star B\star V\)}}


\put(18,39){\makebox(0,8)[bl] {\(S_{1}=H_{1}\star B_{1}\star
V_{2}\)}}

\put(18,35){\makebox(0,8)[bl] {\(S_{2}=H_{2}\star B_{3}\star
V_{2}\)}}

\put(18,31){\makebox(0,8)[bl] {\(S_{3}=H_{3}\star B_{2}\star
V_{1}\)}}

\end{picture}
%
\begin{picture}(60,44)
\put(2,15){\line(0,1){6}} \put(3,15){\line(0,1){6}}

\put(02,15){\line(1,0){17}} \put(02,21){\line(1,0){17}}

\put(3.4,17){\makebox(0,0)[bl]{\(V_{3}\)}}
\put(8,15){\line(0,1){6}}

\put(9,17){\makebox(0,0)[bl]{\(V_{2}\)}}
\put(14,17){\makebox(0,0)[bl]{\(V_{1}\)}}
\put(19,15){\line(0,1){6}}

\put(25,15){\line(1,0){23}} \put(25,21){\line(1,0){23}}
\put(25,15){\line(0,1){6}}
\put(26,17){\makebox(0,0)[bl]{\(B_{2}\)}}
\put(31,15){\line(0,1){6}}
\put(32,17){\makebox(0,0)[bl]{\(B_{3}\)}}
\put(37,15){\line(0,1){6}}
\put(38,17){\makebox(0,0)[bl]{\(B_{1}\)}}
\put(43,17){\makebox(0,0)[bl]{\(B_{4}\)}}
\put(48,15){\line(0,1){6}}
\put(19,25){\line(0,1){18}} \put(25,25){\line(0,1){18}}
\put(19,25){\line(1,0){6}}
\put(20,27){\makebox(0,0)[bl]{\(H_{3}\)}}
\put(20,31){\makebox(0,0)[bl]{\(H_{2}\)}}
\put(19,35){\line(1,0){6}}
\put(20,37){\makebox(0,0)[bl]{\(H_{1}\)}}
\put(19,41){\line(1,0){6}} \put(19,43){\line(1,0){6}}
\put(10,15){\line(0,-1){5}} \put(10,10){\line(1,0){23}}
\put(33,10){\line(0,1){5}} \put(11,11){\makebox(0,0)[bl]{\(2\)}}

\put(10,21){\line(0,1){11}} \put(10,32){\line(1,0){9}}

\put(11,28){\makebox(0,0)[bl]{\(3\)}}

\put(33,21){\line(0,1){11}} \put(33,32){\line(-1,0){8}}

\put(34,28){\makebox(0,0)[bl]{\(2\)}}
\put(9,15){\line(0,-1){6}} \put(9,09){\line(1,0){31}}
\put(40,09){\line(0,1){6}} \put(37,10){\makebox(0,0)[bl]{\(3\)}}
\put(25,18){\line(-1,0){6}} \put(21,14){\makebox(0,0)[bl]{\(1\)}}

\put(28,21){\line(0,1){7}} \put(28,28){\line(-1,0){3}}
\put(29,24){\makebox(0,0)[bl]{\(1\)}}

\put(16,21){\line(0,1){7}} \put(16,28){\line(1,0){3}}
\put(13,24){\makebox(0,0)[bl]{\(1\)}}

\put(9,21){\line(0,1){17}} \put(9,38){\line(1,0){10}}
\put(10,34){\makebox(0,0)[bl]{\(3\)}}
\put(40,21){\line(0,1){17}} \put(40,38){\line(-1,0){15}}
\put(41,33){\makebox(0,0)[bl]{\(3\)}}

\end{picture}
\end{center}

 Fig. 3 illustrates the system quality poset without taking into account
 component compatibility.
  Here poset parameters are: ~\(m=3\), ~\(l=3\).
 The general system quality poset based on
 \(N(S)\) (\(w=1,2,3\)) is depicted in Fig. 4.
 This poset consists of three posets from Fig. 3.

\begin{center}
\begin{picture}(70,86)
\put(04,00){\makebox(0,0)[bl] {Fig. 3.
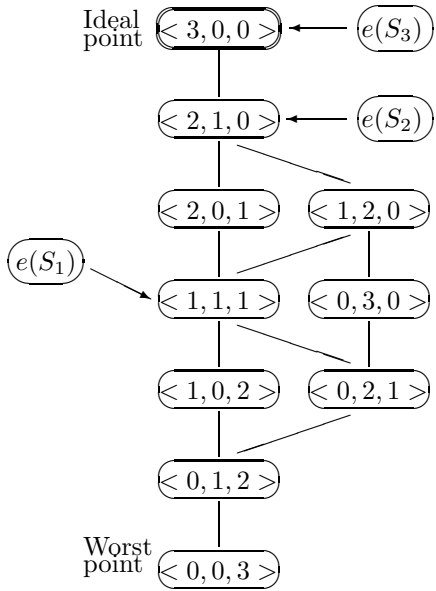
 Poset \(e(S)=(\eta_{1},\eta_{2},\eta_{3})\)}}

\put(010,81){\makebox(0,0)[bl]{Ideal}}
\put(010,78){\makebox(0,0)[bl]{point}}

\put(20,79){\makebox(0,0)[bl]{\(<3,0,0>\) }}
\put(28,81){\oval(16,5)} \put(28,81){\oval(16.5,5.5)}

\put(28,72){\line(0,1){6}}
\put(20,67){\makebox(0,0)[bl]{\(<2,1,0>\)}}
\put(28,69){\oval(16,5)}
\put(47,79){\makebox(0,0)[bl]{\(e(S_{3})\)}}
\put(51.5,81){\oval(10,6)} \put(45,81){\vector(-1,0){7.6}}

\put(47,67){\makebox(0,0)[bl]{\(e(S_{2})\)}}
\put(51.5,69){\oval(10,6)} \put(45,69){\vector(-1,0){8}}

\put(28,60){\line(0,1){6}}
\put(20,55){\makebox(0,0)[bl]{\(<2,0,1>\) }}
\put(28,57){\oval(16,5)}

\put(5,50){\oval(10,6)}
\put(01,48){\makebox(0,0)[bl]{\(e(S_{1})\)}}
\put(11,49){\vector(2,-1){08}}

\put(28,48){\line(0,1){6}}
\put(20,43){\makebox(0,0)[bl]{\(<1,1,1>\) }}
\put(28,45){\oval(16,5)}

\put(28,36){\line(0,1){6}}
\put(20,31){\makebox(0,0)[bl]{\(<1,0,2>\) }}
\put(28,33){\oval(16,5)}

\put(28,24){\line(0,1){6}}
\put(20,19){\makebox(0,0)[bl]{\(<0,1,2>\) }}
\put(28,21){\oval(16,5)}
\put(28,12){\line(0,1){6}}

\put(20,07){\makebox(0,0)[bl]{\(<0,0,3>\) }}
\put(28,09){\oval(16,5)}

\put(010,11){\makebox(0,0)[bl]{Worst}}
\put(010,08){\makebox(0,0)[bl]{point}}
\put(30.5,65.5){\line(3,-1){15}}

\put(40,55){\makebox(0,0)[bl]{\(<1,2,0>\) }}
\put(48,57){\oval(16,5)}
\put(30.5,48.5){\line(3,1){15}}

\put(48,48){\line(0,1){6}}
\put(40,43){\makebox(0,0)[bl]{\(<0,3,0>\) }}
\put(48,45){\oval(16,5)}
\put(30.5,41.5){\line(3,-1){15}}

\put(48,36){\line(0,1){6}}
\put(40,31){\makebox(0,0)[bl]{\(<0,2,1>\) }}
\put(48,33){\oval(16,5)}
\put(45.5,29.5){\line(-3,-1){15}}

\end{picture}
%
\begin{picture}(66.5,63)
\put(4.5,00){\makebox(0,0)[bl] {Fig. 4.
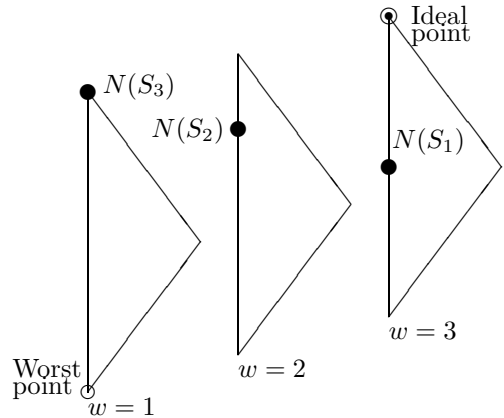
 General system quality poset}}

\put(10,010){\line(0,1){40}} \put(10,010){\line(3,4){15}}
\put(10,050){\line(3,-4){15}}

\put(30,015){\line(0,1){40}} \put(30,015){\line(3,4){15}}
\put(30,055){\line(3,-4){15}}

\put(50,020){\line(0,1){40}} \put(50,020){\line(3,4){15}}
\put(50,060){\line(3,-4){15}}

\put(10,50){\circle*{2}}
\put(12,49){\makebox(0,0)[bl]{\(N(S_{3})\)}}

\put(30,45){\circle*{2}}
\put(18.5,43){\makebox(0,0)[bl]{\(N(S_{2})\)}}

\put(50,40){\circle*{2}}
\put(50.5,41.6){\makebox(0,0)[bl]{\(N(S_{1})\)}}

\put(50,60){\circle{2}} \put(50,60){\circle*{1}}

\put(53,59){\makebox(0,0)[bl]{Ideal}}
\put(53,56){\makebox(0,0)[bl]{point}}

\put(10,07){\makebox(0,0)[bl]{\(w=1\)}}
\put(30,12){\makebox(0,0)[bl]{\(w=2\)}}
\put(50,17){\makebox(0,0)[bl]{\(w=3\)}}

\put(10,10){\circle{1.7}}

\put(00,12){\makebox(0,0)[bl]{Worst}}
\put(00,09){\makebox(0,0)[bl]{point}}

\end{picture}
\end{center}

\subsubsection{Multiset based estimates of system components}

 Fundamentals of multiset theory are presented in
   \cite{knuth98,yager86}.
 Interval estimates based on multisets and their applications in
 combinatorial synthesis have been suggested in
 \cite{lev12a,lev15}.
 Here the following basic scale is used:
 ~ \([1,2,...,l]\) ~ (\(1 \succ 2 \succ 3 \succ ...\)).
 Interval estimate \(e\) for object (alternative)
   \(A\) by scale
  \([1,l]\) is (position representation):~
 \( e(A) = (\eta_{1},...,\eta_{\iota},...,\eta_{l}) \),
 where \(\eta_{\iota}\)
 corresponds to the number of elements at the quality level
  \(\iota\) (\(\iota = \overline{1,l}\)).
 The following conditions are assumed:

 {\it Condition 1}: ~
 \(\sum_{\iota=1}^{l} \eta_{\iota} = \eta\)
 ~ (or \(\left|  e(A)  \right|  = \eta\)).

 {\it Condition 2}: ~
  \((\eta_{\iota} > 0) ~\&~ ( \eta_{\iota+2} > 0 )\) ~
  \( \Longrightarrow \) ~
 \(\eta_{\iota+1} > 0\)  (\(\iota = \overline{1,l-2}\)).

 Presentation of the estimate as multiset is:
 \[e(A) = \{ \overbrace{1,...,1}^{\eta_{1}},\overbrace{2,...2}^{\eta_{2}},
 \overbrace{3,...,3}^{\eta_{3}},...,\overbrace{l,...,l}^{\eta_{l}} \}.\]
 The number of multisets for fixed value of element numbers
  \(\eta\)
 is called coefficient of multiset or multiset number:
 \[ \mu^{l,\eta} =
   \frac{l(l+1)(l+2)... (l+\eta-1) }
   {\eta!} .\]
 This number corresponds to possible number of estimates or
 cardinality
 (without taking into account condition 2).
 In the case of condition 2, the number of the estimates
 is decreased.
 In \cite{lev12a,lev15},
 the following designations
 for assessment problems based on the interval multiset estimates
 are suggested:~
 \(P^{l,\eta}\).

 In the numerical example the following assessment problem
 is used
 \(P^{3,4}\).
 Clearly,
  the basic version of HMMD is based on assessment problem
 \(P^{l,1}\).

 An integrated multiset estimate is described as follows.
 There are  \(n\) initial estimates: ~~
 \[e^{1} = (\eta^{1}_{1},...,\eta^{1}_{\iota},...,\eta^{1}_{l}),
 ...,
 e^{\kappa} = (\eta^{\kappa}_{1},...,\eta^{\kappa}_{\iota},...,\eta^{\kappa}_{l}),
 ...,
 e^{n} = (\eta^{n}_{1},...,\eta^{n}_{\iota},...,\eta^{n}_{l}). \]
 The integrated multiset estimate is:
 \[e^{I} = (\eta^{I}_{1},...,\eta^{I}_{\iota},...,\eta^{I}_{l}),
  ~~~ \eta^{I}_{\iota} = \sum_{\kappa=1}^{n} \eta^{\kappa}_{\iota} ~~ \forall
 \iota = \overline{1,l}. \]
 The following operation is used:
 ~\(\biguplus\): ~
 \(e^{I} = e^{1} \biguplus ... \biguplus e^{\kappa} \biguplus ... \biguplus e^{n}\).

 The vector proximity between two multiset estimates
 ~ \(e(A_{1})\), \(e(A_{2})\) is:
 \[\delta ( e(A_{1}), e(A_{2})) = (\delta^{-}(A_{1},A_{2}),\delta^{+}(A_{1},A_{2})),\]
 where
 (i) \(\delta^{-}\) corresponds to the number of  one-step changes
  (modifications) of quality element
 \(\iota + 1\)
 into quality element
   \(\iota\) (\(\iota = \overline{1,l-1}\))
 (this is improvement);
 (ii) \(\delta^{+}\)
 corresponds to the number of one-step changes (modifications)
 of quality element
 \(\iota\)
 into quality element
  \(\iota+1\) (\(\iota = \overline{1,l-1}\))
  (this is decreasing of quality).
 This description corresponds to modification
 as editing of object (alternative)
  \(A_{1}\) into alternative \(A_{2}\).
 In addition, the following is assumed:~
 \( | \delta ( e(A_{1}), e(A_{2})) | = \max
 \{ | \delta^{-}(A_{1},A_{2}) | , |\delta^{+}(A_{1},A_{2})) | \}\).

 Further, aggregation of estimate (as searching for a median)
 is examined.
 There are a set of estimates
 (as a set of objects/alternatives):
 \[\widehat{E} = \{ e_{1},...,e_{\kappa},...,e_{n}\},\]
 the set of possible estimates is
  \(\widehat{D} \)
 (\( \widehat{E}  \subseteq \widehat{D} \)).
 Aggregation estimate as generalized median is
 \cite{lev11agg,lev15}:
 \[M^{g} =   \arg min_{X \in \widehat{D} } ~~
  \biguplus_{\kappa=1}^{n} ~ |~ \delta (e(X), e_{\kappa}) ~|. \]
 Thus, combinatorial synthesis problem based on multiset estimates
 of DAs is the following:
 \[ \max~ e(S) = M^{g} =
\arg min_{X \in \widehat{D} } ~~
  \biguplus_{\kappa=1}^{n} ~ |~ \delta (e(X), e_{\kappa}) ~|,
%
 ~~~ \max~~ w(S),
%
 ~~~~~s.t. ~~ w(S) \geq 1  .\]
%

\subsection{Aggregation of modular solutions}

 Basic aggregation strategies for modular solutions are considered
 in \cite{lev11agg,lev15}.
 Let   \( \overline{S} = \{S^{1},...,S^{n}\}\) be a set of initial
 modular solutions.
 A general aggregation strategy is targeted to searching for
 consensus/median solution
 \(S^{M}\)
 (this is generalized median)
  for the initial solutions
  \( \overline{S} = \{S^{1},...,S^{n}\}\):
 \[ S^{M} = \arg ~ min_{X \in \overline{S}}  ~
 ( \sum_{i=1}^{n} ~ \rho (X, S^{i})  ),\]
  where
 \(\rho (X, Y)\) is a proximity between two solutions
   \(X,Y \in \overline{S} \).
 This problem (searching for the generalized median)
 is often NP-hard.
 It may be reasonable to use simplified (approximate)
 strategies, for example:
  (a) selection of solution from the set of initial solutions
  (i.e., set median),
  (b) extension strategy,
  (c) compressed strategy.
 The last two strategies are as follows:


 {\bf 1.} {\it Extension strategy}:
 ~{\it 1.1.} design of a 'kernel'
 for the initial solutions
 (substructure or an extended substructure),
 ~{\it 1.2.} generation of some additional elements
 for possible inclusion into the 'kernel',
 ~{\it 1.3.} selection of the additional elements while
 taking into account their 'profit'
 and resource requirements (e.g., cost)
 (here basic knapsack problem can be used).


  {\bf 2.} {\it Compression strategy}:
 ~{\it 2.1.} design a super structure for the initial solutions,
 ~{\it 2.2.}
 generation of the superstructure elements as possible candidates
 for deletion,
 ~{\it 2.3.}
 selection of the elements for deletion from the superstructure
 while taking into account their 'profit'
 resource requirements (e.g., cost)
 (here knapsack problem with minimization of the objective function
 can be used).

 In the paper, extension aggregation strategy is used.

\section{Geological exploration}

 In this section,
 combinatorial planning of geological exploration
 is examined as a numerical example for
 for oil and gas fields of Yamal peninsula
 \cite{maximov79}
 (Fig. 5).
 The general plan involves five parts:
  ~\(S = A^{1} \star A^{2} \star A^{3} \star A^{4} \star A^{5}\),
 where
  \(A^{1}\) corresponds to field Kharosovey,
  \(A^{2}\) corresponds to field Arkticheskoe,
  \(A^{3}\) corresponds to field Neitinskoe,
  \(A^{4}\) corresponds to Kruzensternskoe,
  \(A^{5}\) corresponds to field Bovanenkovskoe.

\begin{center}
\begin{picture}(81,40)
\put(13,00){\makebox(0,0)[bl] {Fig. 5.
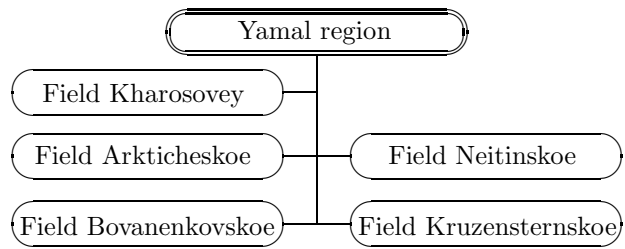
 Oil-gas fields (Yamal region)}}

\put(30,33.9){\makebox(0,0)[bl]{Yamal region}}

\put(40.5,35.5){\oval(40,06)} \put(40.5,35.5){\oval(39,05)}

\put(40.5,32.5){\line(0,-1){22.5}}

\put(36,27.5){\line(1,0){4.5}}

\put(36,10){\line(1,0){9}} \put(36,19){\line(1,0){9}}

\put(18,27.5){\oval(36,06)}

\put(04,25.5){\makebox(0,0)[bl]{Field Kharosovey}}

\put(18,19){\oval(36,06)}

\put(03,18){\makebox(0,0)[bl]{Field Arkticheskoe}}

\put(63,19){\oval(36,06)}

\put(50,18){\makebox(0,0)[bl]{Field Neitinskoe}}

\put(18,10){\oval(36,06)}

\put(01,08.5){\makebox(0,0)[bl]{Field Bovanenkovskoe}}
\put(63,10){\oval(36,06)}

\put(46,08.5){\makebox(0,0)[bl]{Field Kruzensternskoe}}

\end{picture}
\end{center}

 The solving scheme consists of two stages:

 {\it Stage 1.}
 Hierarchical combinatorial construction of
 the exploration plan for oil and gas fields
 (here only two oil and gas fields are described).

 {\it Stage 2.} Composition of the general exploration
 plan for region.

\subsection{Problem formulation}

 The following four-layer hierarchy of geological objects is considered:
 (a) productive stratum (reservoir) (bottom hierarchical level);
 (b) bore hole as a group of productive stratums (reservoirs);
 (c) oil and gas field;
 (d) group of oil and gas fields (region).
 The following assessment parameters (attributes) are used:

 {\it 1.} parameter of reservoir existence
 ('3' corresponds to existence of reservoir,
 '2' corresponds to prospective geological position
  (horizon) in the field,
  '1' corresponds to prospective geological position
  (horizon) in the traprock);

  {\it 2.} cover of thickness, m;

  {\it 3.} type of fluid, i.e.,  classification factor:
   (i) gas, (ii) gas and condensate (condensed fluid),
   (iii) oil;

  {\it 4.} volume of geological reserves or resources
  (gas  - million cubic meters,
  oil -  thousand tonnes);

  {\it 5.} production rate of work wellsite
   (cubic metes per 24 hours);

  {\it 6.} complexity of geological situation
  ('1' - simple,
   '2' - complex,
   '3' - very complex);

  {\it 7.} reliability (risk) to obtain the results
  ([0...100]);

  {\it 8.} validity (adequacy) of assessment
 of geological reserves (i.e., oil/gas/condensate in place,
   probable reserves)
   (\(C_{1}\) - \(20 \%\),
 \(C_{2}\) - \(50 \%\),
 \(C_{3}\) - \(80 \%\),
  etc.);

 {\it 9.} proximity to technological base (gas-oil pipeline, km).

 Eight  DAs are examined for
 each geological object (as stratum) (Table 1)
 (the corresponding bottom index is used for the designation).

\begin{center}
\begin{small}
 {\bf Table 1.} Design alternatives for geological exploration\\
\begin{tabular}{|c|c|l |}
\hline
 & Notation & Content of geological exploration \\

\hline

 1.&\(X_{1}\)&conservation\\

 2.&\(X_{2}\)& appraisal work \\ 

 3.&\(X_{3}\) &independent production time (gas)\\

 4.&\(X_{4}\)&independent production time (oil)\\

 5.&\(X_{5}\)&independent production time (oil and gas)\\

 6.&\(X_{6}\)&joint independent production time (gas)\\

 7.&\(X_{7}\)&joint independent production time (oil)\\

 8.&\(X_{8}\)&joint independent production time (oil and gas)\\

\hline
\end{tabular}
\end{small}
\end{center}

 Further,
 a subset of alternative actions (DAs) for each geological object
 (stratum) at bottom layer of the system model
 (i.e., productive stratum)
 is selected (from initial eight basic DAs) (expert judgment).
 This a preliminary selection at the bottom layer of the problem.
 At the next step,
 the selected DAs are used as a basis for composition of composite
 DAs for more higher layer of the problem
 (i.e., for group of geological objects as bore holes,
  and for fields)

 Each strategy component
 (geological object, group of objects, strategy)
 is noted  by symbol
 (the level of effectiveness of priority
 ~\(\iota\)~ is pointed out for each components
 in  parentheses).
 It is assumed that experts have their skills for the following:
 (1) selection of DAs for each geological object,
 (2) ranking of DAs for each geological object,
 (3) assessment of compatibility among DAs
  (by an ordinal scale).

 The illustrative hierarchical model of oil and gas field is
 depicted in Fig. 6.

\begin{center}
\begin{picture}(110,87)
\put(12,00){\makebox(0,0)[bl]{Fig. 6.
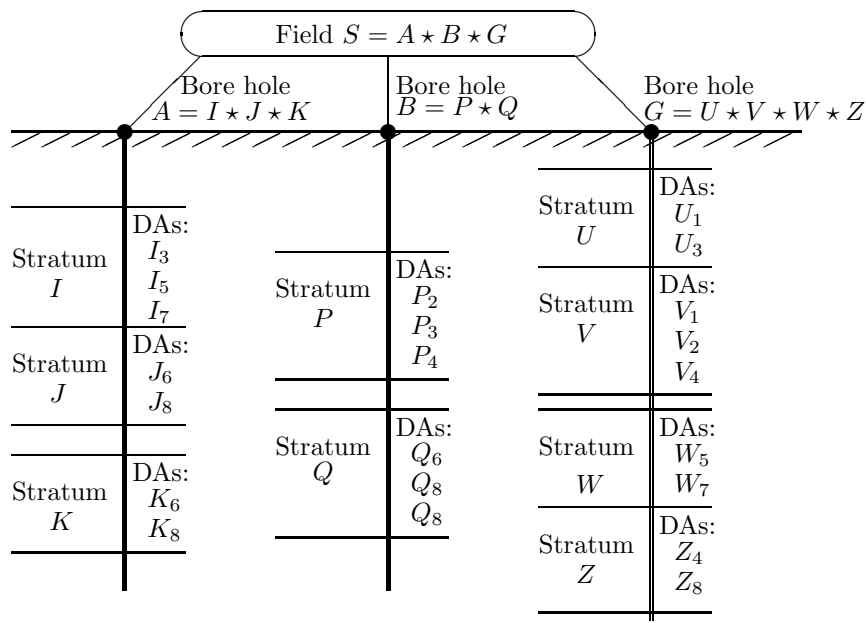
 Illustration for hierarchical model of field}}

\put(50,83){\oval(55,06)}

\put(35,81.75){\makebox(0,0)[bl]{Field \(S = A\star B \star G\)}}

\put(15,70){\circle*{2}} \put(15,70){\line(1,1){10}}

\put(22.4,75){\makebox(0,0)[bl]{Bore hole}}
\put(18.5,71.5){\makebox(0,0)[bl]{\(A = I\star J \star K \)}}

\put(15.2,70){\line(0,-1){61}} \put(14.8,70){\line(0,-1){61}}

\put(00,60){\line(1,0){23}}

\put(00,52){\makebox(0,0)[bl]{Stratum}}
\put(05,48){\makebox(0,0)[bl]{\(I\)}}

\put(16,56.5){\makebox(0,0)[bl]{DAs:}}
\put(18,52.5){\makebox(0,0)[bl]{\(I_{3}\)}}
\put(18,48.5){\makebox(0,0)[bl]{\(I_{5}\)}}
\put(18,44.5){\makebox(0,0)[bl]{\(I_{7}\)}}

\put(00,44){\line(1,0){23}}

\put(00,38){\makebox(0,0)[bl]{Stratum}}
\put(05,34){\makebox(0,0)[bl]{\(J\)}}

\put(16,40.5){\makebox(0,0)[bl]{DAs:}}
\put(18,36.5){\makebox(0,0)[bl]{\(J_{6}\)}}
\put(18,32.5){\makebox(0,0)[bl]{\(J_{8}\)}}

\put(00,31){\line(1,0){23}}

\put(00,27){\line(1,0){23}}

\put(00,21){\makebox(0,0)[bl]{Stratum}}
\put(05,17){\makebox(0,0)[bl]{\(K\)}}

\put(16,23.5){\makebox(0,0)[bl]{DAs:}}
\put(18,19.5){\makebox(0,0)[bl]{\(K_{6}\)}}
\put(18,15.5){\makebox(0,0)[bl]{\(K_{8}\)}}

\put(00,14){\line(1,0){23}}

\put(50,70){\circle*{2}} \put(50,70){\line(0,1){10}}

\put(51,75){\makebox(0,0)[bl]{Bore hole}}
\put(51,71.5){\makebox(0,0)[bl]{\(B = P\star Q\)}}

\put(50.2,70){\line(0,-1){61}} \put(49.8,70){\line(0,-1){61}}

\put(35,54){\line(1,0){23}}

\put(35,48){\makebox(0,0)[bl]{Stratum}}
\put(40,44){\makebox(0,0)[bl]{\(P\)}}

\put(51,50.5){\makebox(0,0)[bl]{DAs:}}
\put(53,46.5){\makebox(0,0)[bl]{\(P_{2}\)}}
\put(53,42.5){\makebox(0,0)[bl]{\(P_{3}\)}}
\put(53,38.5){\makebox(0,0)[bl]{\(P_{4}\)}}

\put(35,37){\line(1,0){23}}

\put(35,33){\line(1,0){23}}

\put(35,27){\makebox(0,0)[bl]{Stratum}}
\put(40,23){\makebox(0,0)[bl]{\(Q\)}}

\put(51,29.5){\makebox(0,0)[bl]{DAs:}}
\put(53,25.5){\makebox(0,0)[bl]{\(Q_{6}\)}}
\put(53,21.5){\makebox(0,0)[bl]{\(Q_{8}\)}}
\put(53,17.5){\makebox(0,0)[bl]{\(Q_{8}\)}}

\put(35,16){\line(1,0){23}}

\put(85,70){\circle*{2}} \put(85,70){\line(-1,1){10}}

\put(84,75){\makebox(0,0)[bl]{Bore hole}}
\put(84,71.5){\makebox(0,0)[bl]{\(G = U\star V \star W \star Z\)}}

\put(85.2,70){\line(0,-1){65}} \put(84.8,70){\line(0,-1){65}}

\put(70,65){\line(1,0){23}}

\put(70,59){\makebox(0,0)[bl]{Stratum}}
\put(75,55){\makebox(0,0)[bl]{\(U\)}}

\put(86,61.5){\makebox(0,0)[bl]{DAs:}}
\put(88,57.5){\makebox(0,0)[bl]{\(U_{1}\)}}
\put(88,53.5){\makebox(0,0)[bl]{\(U_{3}\)}}

\put(70,52){\line(1,0){23}}


\put(70,46){\makebox(0,0)[bl]{Stratum}}
\put(75,42){\makebox(0,0)[bl]{\(V\)}}

\put(86,48.5){\makebox(0,0)[bl]{DAs:}}
\put(88,44.5){\makebox(0,0)[bl]{\(V_{1}\)}}
\put(88,40.5){\makebox(0,0)[bl]{\(V_{2}\)}}
\put(88,36.5){\makebox(0,0)[bl]{\(V_{4}\)}}

\put(70,35){\line(1,0){23}}

\put(70,33){\line(1,0){23}}

\put(70,27){\makebox(0,0)[bl]{Stratum}}
\put(75,22){\makebox(0,0)[bl]{\(W\)}}

\put(86,29.5){\makebox(0,0)[bl]{DAs:}}
\put(88,25.5){\makebox(0,0)[bl]{\(W_{5}\)}}
\put(88,21.5){\makebox(0,0)[bl]{\(W_{7}\)}}

\put(70,20){\line(1,0){23}}


\put(70,14){\makebox(0,0)[bl]{Stratum}}
\put(75,10){\makebox(0,0)[bl]{\(Z\)}}

\put(86,16.5){\makebox(0,0)[bl]{DAs:}}
\put(88,12.5){\makebox(0,0)[bl]{\(Z_{4}\)}}
\put(88,08.5){\makebox(0,0)[bl]{\(Z_{8}\)}}

\put(70,06){\line(1,0){23}}

\put(00,70){\line(1,0){105}}

\put(00,68){\line(2,1){4}} \put(04,68){\line(2,1){4}}
\put(08,68){\line(2,1){4}} \put(12,68){\line(2,1){4}}
\put(16,68){\line(2,1){4}} \put(20,68){\line(2,1){4}}
\put(24,68){\line(2,1){4}} \put(28,68){\line(2,1){4}}
\put(32,68){\line(2,1){4}} \put(36,68){\line(2,1){4}}
\put(40,68){\line(2,1){4}} \put(44,68){\line(2,1){4}}
\put(48,68){\line(2,1){4}} \put(52,68){\line(2,1){4}}
\put(56,68){\line(2,1){4}} \put(60,68){\line(2,1){4}}
\put(64,68){\line(2,1){4}} \put(68,68){\line(2,1){4}}
\put(72,68){\line(2,1){4}} \put(76,68){\line(2,1){4}}
\put(80,68){\line(2,1){4}} \put(84,68){\line(2,1){4}}
\put(88,68){\line(2,1){4}} \put(92,68){\line(2,1){4}}
\put(96,68){\line(2,1){4}} \put(100,68){\line(2,1){4}}

\end{picture}
\end{center}

\subsection{Examples for oil-gas fields}

 The modular exploration strategy for field Arkticheskoe
 is shown in Fig. 7.
 Table 2 contains compatibility factors for
 the strategy elements
 (here 'C5+v' corresponds to level of
 hydrocarbon in gas as 'C5' and more).

\begin{center}
\begin{picture}(102,82)
\put(22,00){\makebox(0,0)[bl]{Fig. 7.
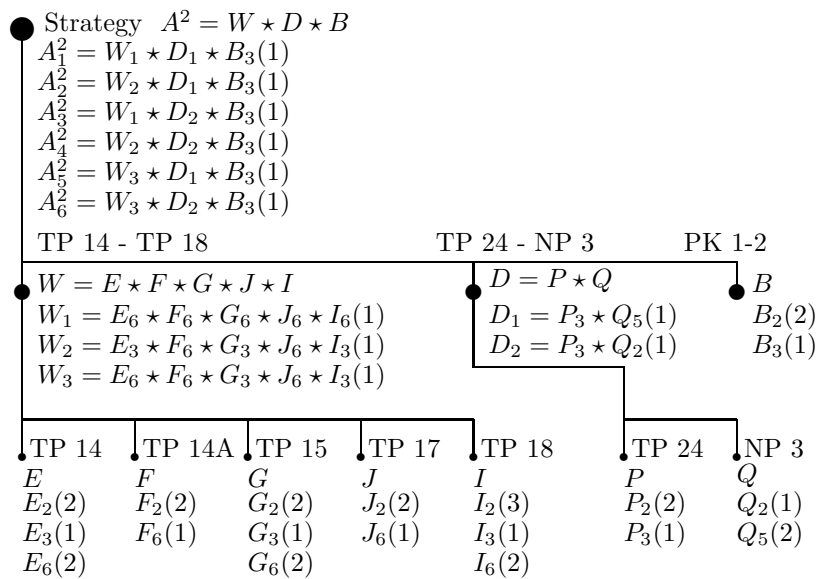
 Strategy for field Arkticheskoe}}

\put(00,78){\circle*{3}}
\put(03,77){\makebox(0,0)[bl]{Strategy~~\(A^{2}=W\star D\star
 B\)}}

\put(02,73){\makebox(0,0)[bl]{\(A^{2}_{1}= W_{1}\star D_{1}\star
B_{3}(1)\)}}

\put(02,69){\makebox(0,0)[bl]{\(A^{2}_{2}= W_{2}\star D_{1}\star
B_{3}(1)\)}}

\put(02,65){\makebox(0,0)[bl]{\(A^{2}_{3}= W_{1}\star D_{2}\star
B_{3}(1)\)}}

\put(02,61){\makebox(0,0)[bl]{\(A^{2}_{4}= W_{2}\star D_{2}\star
B_{3}(1)\)}}

\put(02,57){\makebox(0,0)[bl]{\(A^{2}_{5}= W_{3}\star D_{1}\star
B_{3}(1)\)}}

\put(02,53){\makebox(0,0)[bl]{\(A^{2}_{6}= W_{3}\star D_{2}\star
B_{3}(1)\)}}


\put(02,48.5){\makebox(0,0)[bl]{TP 14 - TP 18}}

\put(55,48.5){\makebox(0,0)[bl]{TP 24 - NP 3}}

\put(88,48.5){\makebox(0,0)[bl]{PK 1-2}}

\put(00,48){\line(0,1){30}} \put(00,47){\line(1,0){95}}

\put(95,43){\line(0,1){04}} \put(95,43){\circle*{2}}

\put(97,43){\makebox(0,0)[bl]{\(B\)}}
\put(97,38){\makebox(0,0)[bl]{\(B_{2}(2)\)}}
\put(97,34){\makebox(0,0)[bl]{\(B_{3}(1)\)}}

\put(60,43){\line(0,1){04}} \put(60,43){\line(0,-1){10}}

\put(60,33){\line(1,0){20}}

\put(80,33){\line(0,-1){12}}

\put(60,43){\circle*{2}}

\put(62,43){\makebox(0,0)[bl]{\(D=P\star Q\)}}
\put(62,38){\makebox(0,0)[bl]{\(D_{1}=P_{3}\star Q_{5}(1)\)}}
\put(62,34){\makebox(0,0)[bl]{\(D_{2}=P_{3}\star Q_{2}(1)\)}}

\put(00,43){\line(0,1){05}} \put(00,43){\circle*{2}}

\put(00,26){\line(0,1){21}}

\put(02,43){\makebox(0,0)[bl]{\(W=E\star F\star G\star J \star I
\)}}

\put(02,38){\makebox(0,0)[bl]{\(W_{1}=E_{6}\star F_{6}\star
 G_{6}\star J_{6}\star I_{6}(1)\)}}

\put(02,34){\makebox(0,0)[bl]{\(W_{2}=E_{3}\star F_{6}\star G_{3}
\star J_{6}\star I_{3}(1)\)}}

\put(02,30){\makebox(0,0)[bl]{\(W_{3}=E_{6}\star F_{6}\star G_{3}
\star J_{6}\star I_{3}(1)\)}}

\put(00,26){\line(1,0){60}} \put(80,26){\line(1,0){15}}

\put(00,21){\line(0,1){05}} \put(15,21){\line(0,1){05}}
\put(30,21){\line(0,1){05}} \put(45,21){\line(0,1){05}}
\put(60,21){\line(0,1){05}}

\put(80,21){\line(0,1){05}} \put(95,21){\line(0,1){05}}

\put(00,21){\circle*{1}} \put(15,21){\circle*{1}}
\put(30,21){\circle*{1}} \put(45,21){\circle*{1}}
\put(60,21){\circle*{1}}

\put(80,21){\circle*{1}} \put(95,21){\circle*{1}}
\put(01,21){\makebox(0,0)[bl]{TP 14}}
\put(16,21){\makebox(0,0)[bl]{TP 14A}}
\put(31,21){\makebox(0,0)[bl]{TP 15}}
\put(46,21){\makebox(0,0)[bl]{TP 17}}
\put(61,21){\makebox(0,0)[bl]{TP 18}}
\put(81,21){\makebox(0,0)[bl]{TP 24}}
\put(96,21){\makebox(0,0)[bl]{NP 3}}

\put(95,17){\makebox(0,0)[bl]{\(Q\)}}
\put(95,13){\makebox(0,0)[bl]{\(Q_{2}(1)\)}}
\put(95,09){\makebox(0,0)[bl]{\(Q_{5}(2)\)}}

\put(80,17){\makebox(0,0)[bl]{\(P\)}}
\put(80,13){\makebox(0,0)[bl]{\(P_{2}(2)\)}}
\put(80,09){\makebox(0,0)[bl]{\(P_{3}(1)\)}}

\put(60,17){\makebox(0,0)[bl]{\(I\)}}
\put(60,13){\makebox(0,0)[bl]{\(I_{2}(3)\)}}
\put(60,09){\makebox(0,0)[bl]{\(I_{3}(1)\)}}
\put(60,05){\makebox(0,0)[bl]{\(I_{6}(2)\)}}

\put(45,17){\makebox(0,0)[bl]{\(J\)}}
\put(45,13){\makebox(0,0)[bl]{\(J_{2}(2)\)}}
\put(45,09){\makebox(0,0)[bl]{\(J_{6}(1)\)}}

\put(30,17){\makebox(0,0)[bl]{\(G\)}}
\put(30,13){\makebox(0,0)[bl]{\(G_{2}(2)\)}}
\put(30,09){\makebox(0,0)[bl]{\(G_{3}(1)\)}}
\put(30,05){\makebox(0,0)[bl]{\(G_{6}(2)\)}}

\put(15,17){\makebox(0,0)[bl]{\(F\)}}
\put(15,13){\makebox(0,0)[bl]{\(F_{2}(2)\)}}
\put(15,09){\makebox(0,0)[bl]{\(F_{6}(1)\)}}

\put(00,17){\makebox(0,0)[bl]{\(E\)}}
\put(00,13){\makebox(0,0)[bl]{\(E_{2}(2)\)}}
\put(00,09){\makebox(0,0)[bl]{\(E_{3}(1)\)}}
\put(00,05){\makebox(0,0)[bl]{\(E_{6}(2)\)}}

\end{picture}
\end{center}

\newpage
\begin{center}
\begin{small}
 {\bf Table 2.} Compatibility factors for DAs pair \\
\begin{tabular}{|c|l|l |}
\hline
  &DA \& DA & Factors \\

\hline
 1.& TP 14 E \& TP 14A  F & Geological reserves, proximity, 'C5+v'\\

 2.& TP 14  E \& TP 15 G & Geological reserves, proximity, 'C5+v'  \\

 3.& TP 14 E \& TP 17 J & Geological reserves, proximity\\

 4.& TP 14 E \& TP 18 I & Geological reserves, proximity  \\

 5.& TP 14A  F \& TP 15  G& Geological reserves, proximity\\

 6.& TP 14A F \& TP 17  J& Geological reserves, proximity \\

 7.& TP 14A F \& T 18 I & Proximity   \\

 8.& TP 15 G \& TP 17 J & Geological reserves, proximity, 'C5+v'  \\

 9.& TP 15 G \& TP 18 I & Geological reserves, proximity, 'C5+v'  \\

 10.& TP 17 J \& TP 18  I & Geological reserves, proximity, 'C5+v' \\

 11.& TP 24 P \& NP 3 Q & Proximity, 'C5+v' \\

\hline
\end{tabular}
\end{small}
\end{center}

 Compatibility estimates between DAs (expert judgment)
 are contained in Table 3 and Table 4.
 Composite DAs for group of geological objects
 are obtained for
 TP 14 - TP 18 (W),  TP 24 - NP3 (D).
 Thus, 6 versions of exploration strategy (field Arkticheskoe)
  are obtained:

 \(A^{2}_{1}= W_{1}\star D_{1}\star B_{3}(1)\),
 \(A^{2}_{2}= W_{2}\star D_{1}\star B_{3}(1)\),
 \(A^{2}_{3}= W_{1}\star D_{2}\star B_{3}(1)\),

 \(A^{2}_{4}= W_{2}\star D_{2}\star B_{3}(1)\),
 \(A^{2}_{5}= W_{3}\star D_{1}\star B_{3}(1)\),
 \(A^{2}_{6}= W_{3}\star D_{2}\star B_{3}(1)\).

 Table 6 contains some examples of bottlenecks and possible
 improvement operations.
 Fig. 8 depicts quality of composite DAs for component \(W\).

 The exploration strategy for field Kruzensternskoe is shown in
 Fig. 9.
 Table 7 contains compatibility factors for strategy elements.
 The compatibility estimates among DAs (expert judgment)
 are presented in Table 8, Table 9.
 Composite DAs for
  \(B\) and \(H\) are presented in Table 10.
 The obtained two solutions for field Kruzensternskoe are:
 \(A^{4}_{1}= B_{1}\star H_{1}\),
 \(A^{4}_{2}= B_{2}\star H_{2}\).
 Fig. 10 illustrates quality of composite solutions for component
 \(H\).

\begin{center}
\begin{small}
 {\bf Table 3.}  Compatibility for DAs (groups TP 14 - TP 18, part \(W\))\\
\begin{tabular}{|c| ccccc ccccc |}
\hline
 &\(F_{2}\)&\(F_{6}\)&\(G_{2}\)&\(G_{3}\)&\(G_{6}\)
 &\(J_{2}\)&\(J_{6}\)&\(I_{2}\)&\(I_{3}\)&\(I_{6}\)\\

\hline

 \(E_{2}\)
 &\(2\)&\(3\)&\(2\)&\(3\)&\(4\)
 &\(1\)&\(0\)&\(2\)&\(3\)&\(3\)\\

 \(E_{3}\)
 &\(3\)&\(4\)&\(2\)&\(4\)&\(4\)
 &\(1\)&\(2\)&\(3\)&\(2\)&\(2\)\\

 \(E_{6}\)
 &\(3\)&\(4\)&\(3\)&\(4\)&\(4\)
 &\(1\)&\(4\)&\(3\)&\(3\)&\(4\)\\

 \(F_{2}\)&\(\)&\(\)
 &\(2\)&\(3\)&\(4\)&\(2\)&\(3\)&\(2\)&\(3\)&\(2\)\\

  \(F_{6}\)&\(\)&\(\)
 &\(1\)&\(3\)&\(4\)&\(3\)&\(3\)&\(2\)&\(3\)&\(1\)\\

 \(G_{2}\)&\(\)&\(\)&\(\)&\(\)&\(\)
 &\(2\)&\(3\)&\(2\)&\(1\)&\(3\)\\

 \(G_{3}\)&\(\)&\(\)&\(\)&\(\)&\(\)
 &\(2\)&\(4\)&\(2\)&\(3\)&\(1\)\\

 \(G_{6}\)&\(\)&\(\)&\(\)&\(\)&\(\)
 &\(2\)&\(4\)&\(2\)&\(3\)&\(1\)\\

 \(J_{2}\)&\(\)&\(\)&\(\)&\(\)&\(\)&\(\)&\(\)
 &\(2\)&\(1\)&\(1\)\\

 \(J_{6}\)&\(\)&\(\)&\(\)&\(\)&\(\)&\(\)&\(\)
 &\(1\)&\(3\)&\(4\)\\

\hline
\end{tabular}
\end{small}
\end{center}

\begin{center}
\begin{small}
 {\bf Table 4.}  Compatibility for DAs (groups TP 24 - NP 3, part \(D\))\\
\begin{tabular}{|c| cc |}
\hline
 &\(Q_{2}\)&\(Q_{5}\)\\

\hline

 \(P_{2}\)&\(2\)&\(3\)\\

 \(P_{3}\)&\(3\)&\(4\)\\

\hline
\end{tabular}
\end{small}
\end{center}

\begin{center}
\begin{small}
 {\bf Table 5.} Composite DAs \\
\begin{tabular}{|l|c| }
\hline
 Intermediate composite DAs&\(N\)\\

\hline

 \(D_{1}=P_{3}\star Q_{5}\)
 &\(4;1,1,0\)\\

 \(D_{2}=P_{3}\star Q_{2}\)
 &\(3;2,0,0\)\\

 \(W_{1}=E_{6}\star F_{6}\star G_{6}\star J_{6}\star I_{6}\)
 &\(4;2,3,0\)\\

 \(W_{2}=E_{3}\star F_{6}\star G_{3}\star J_{6}\star I_{3}\)
 &\(2;5,0,0\)\\

 \(W_{3}=E_{6}\star F_{6}\star G_{3}\star J_{6}\star I_{3}\)
 &\(3;4,1,0\)\\

\hline
\end{tabular}
\end{small}
\end{center}

\newpage
\begin{center}
\begin{small}
 {\bf Table 6.} Bottlenecks and improvement operations\\
\begin{tabular}{|l|l|ccc|c| }
\hline
  &Intermediate composite DAs&Bottlenecks:&DAs&IC&Improvement\\
    &&&&&operation\\
    &            &            &   &  & \(w/r\)\\

\hline
 1.&\(D_{1}=P_{3}\star Q_{5}\)
   &  &  \(Q_{5}\) &  &\(2 \Rightarrow 1\) \\

 2.&\(D_{2}=P_{3}\star Q_{2}\)
    &  &  &  \((P_{3},Q_{2})\) &\(3 \Rightarrow 4\)  \\

 3.&\(W_{1}= E_{6}\star F_{6} \star G_{6} \star J_{6}\star I_{6}\)
   &  &  \(E_{6}\) &  &\(2 \Rightarrow 1\) \\

 4.&\(W_{1}= E_{6}\star F_{6} \star G_{6} \star J_{6}\star I_{6}\)
   &  &  \(G_{6}\) &  &\(2 \Rightarrow 1\) \\

 5.&\(W_{1}= E_{6}\star F_{6} \star G_{6} \star J_{6}\star I_{6}\)
   &  &  \(I_{6}\) & &\(2 \Rightarrow 1\) \\

 6.&\(W_{3}= E_{6}\star F_{6} \star G_{3} \star J_{6}\star I_{3}\)
   &  &  \(E_{6}\) &  &\(2 \Rightarrow 1\)  \\

\hline
\end{tabular}
\end{small}
\end{center}

\begin{center}
\begin{picture}(85,68)
\put(14,00){\makebox(0,0)[bl] {Fig. 8.
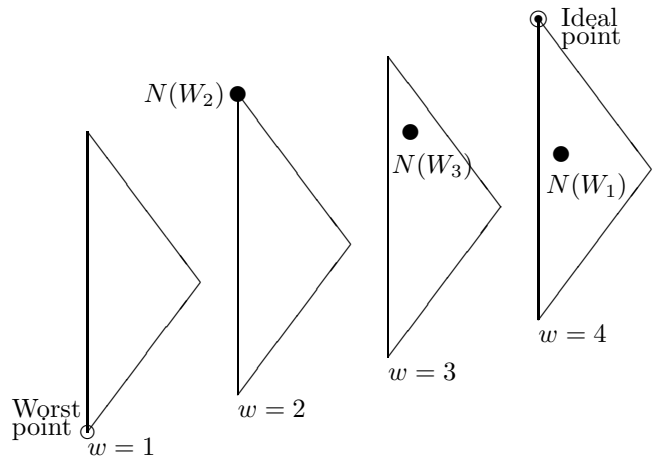
 System quality poset for  \(N(W)\)}}

\put(10,010){\line(0,1){40}} \put(10,010){\line(3,4){15}}
\put(10,050){\line(3,-4){15}}

\put(30,015){\line(0,1){40}} \put(30,015){\line(3,4){15}}
\put(30,055){\line(3,-4){15}}

\put(50,020){\line(0,1){40}} \put(50,020){\line(3,4){15}}
\put(50,060){\line(3,-4){15}}

\put(70,025){\line(0,1){40}} \put(70,025){\line(3,4){15}}
\put(70,065){\line(3,-4){15}}

\put(30,55){\circle*{2}}
\put(17.5,53){\makebox(0,0)[bl]{\(N(W_{2})\)}}

\put(53,50){\circle*{2}}
\put(50.5,43.6){\makebox(0,0)[bl]{\(N(W_{3})\)}}

\put(73,47){\circle*{2}}
\put(71,41){\makebox(0,0)[bl]{\(N(W_{1})\)}}

\put(70,65){\circle{2}} \put(70,65){\circle*{1}}


\put(73,64){\makebox(0,0)[bl]{Ideal}}
\put(73,61){\makebox(0,0)[bl]{point}}

\put(10,07){\makebox(0,0)[bl]{\(w=1\)}}
\put(30,12){\makebox(0,0)[bl]{\(w=2\)}}
\put(50,17){\makebox(0,0)[bl]{\(w=3\)}}
\put(70,22){\makebox(0,0)[bl]{\(w=4\)}}

\put(10,10){\circle{1.7}}

\put(0,12){\makebox(0,0)[bl]{Worst}}
\put(0,09){\makebox(0,0)[bl]{point}}

\end{picture}
\end{center}

\begin{center}
\begin{picture}(93,89)
\put(015,00){\makebox(0,0)[bl]{Fig. 9.
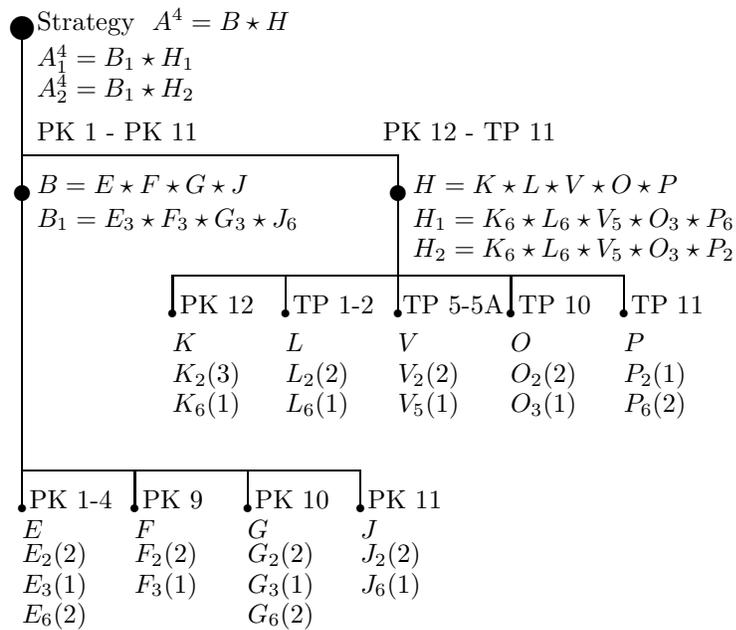
  Strategy for field Kruzensternskoe}}

\put(00,85){\circle*{3}}

\put(02,84){\makebox(0,0)[bl]{Strategy~~\(A^{4}=B\star H\)}}
\put(02,79){\makebox(0,0)[bl]{\(A^{4}_{1}= B_{1}\star H_{1}\)}}
\put(02,75){\makebox(0,0)[bl]{\(A^{4}_{2}= B_{1}\star H_{2}\)}}
\put(02,70){\makebox(0,0)[bl]{PK 1 - PK 11}}
\put(48,70){\makebox(0,0)[bl]{PK 12 - TP 11}}

\put(00,68){\line(0,1){17}} \put(00,68){\line(1,0){50}}

\put(50,63){\line(0,1){05}}

\put(50,63){\line(0,-1){16}}

\put(50,63){\circle*{2}}

\put(52,63){\makebox(0,0)[bl]{\(H=K\star L\star V\star O\star
P\)}}

\put(52,58){\makebox(0,0)[bl]{\(H_{1}=K_{6}\star L_{6} \star V_{5}
\star O_{3} \star P_{6}\)}}

\put(52,54){\makebox(0,0)[bl]{\(H_{2}=K_{6}\star L_{6} \star V_{5}
\star O_{3} \star P_{2}\)}}

\put(00,63){\line(0,1){05}} \put(00,63){\circle*{2}}

\put(00,22){\line(0,1){41}}

\put(02,63){\makebox(0,0)[bl]{\(B=E\star F\star G\star J\)}}

\put(02,58){\makebox(0,0)[bl]{\(B_{1}=E_{3}\star F_{3}\star
G_{3}\star J_{6}\)}}

\put(00,26){\line(1,0){45}}

\put(00,21){\line(0,1){05}} \put(15,21){\line(0,1){05}}
\put(30,21){\line(0,1){05}} \put(45,21){\line(0,1){05}}

\put(00,21){\circle*{1}} \put(15,21){\circle*{1}}
\put(30,21){\circle*{1}} \put(45,21){\circle*{1}}

\put(01,21){\makebox(0,0)[bl]{PK 1-4}}
\put(16,21){\makebox(0,0)[bl]{PK 9}}
\put(31,21){\makebox(0,0)[bl]{PK 10}}
\put(46,21){\makebox(0,0)[bl]{PK 11}}

\put(20,52){\line(1,0){60}}

\put(20,47){\line(0,1){05}} \put(35,47){\line(0,1){05}}
\put(50,47){\line(0,1){05}} \put(65,47){\line(0,1){05}}
\put(80,47){\line(0,1){05}}

\put(20,47){\circle*{1}} \put(35,47){\circle*{1}}
\put(50,47){\circle*{1}} \put(65,47){\circle*{1}}
\put(80,47){\circle*{1}}

\put(21,47){\makebox(0,0)[bl]{PK 12}}
\put(36,47){\makebox(0,0)[bl]{TP 1-2}}
\put(50.6,47){\makebox(0,0)[bl]{TP 5-5A}}
\put(66,47){\makebox(0,0)[bl]{TP 10}}
\put(81,47){\makebox(0,0)[bl]{TP 11}}

\put(80,42){\makebox(0,0)[bl]{\(P\)}}
\put(80,37){\makebox(0,0)[bl]{\(P_{2}(1)\)}}
\put(80,33){\makebox(0,0)[bl]{\(P_{6}(2)\)}}

\put(65,42){\makebox(0,0)[bl]{\(O\)}}
\put(65,37){\makebox(0,0)[bl]{\(O_{2}(2)\)}}
\put(65,33){\makebox(0,0)[bl]{\(O_{3}(1)\)}}

\put(50,42){\makebox(0,0)[bl]{\(V\)}}
\put(50,37){\makebox(0,0)[bl]{\(V_{2}(2)\)}}
\put(50,33){\makebox(0,0)[bl]{\(V_{5}(1)\)}}

\put(35,42){\makebox(0,0)[bl]{\(L\)}}
\put(35,37){\makebox(0,0)[bl]{\(L_{2}(2)\)}}
\put(35,33){\makebox(0,0)[bl]{\(L_{6}(1)\)}}

\put(20,42){\makebox(0,0)[bl]{\(K\)}}
\put(20,37){\makebox(0,0)[bl]{\(K_{2}(3)\)}}
\put(20,33){\makebox(0,0)[bl]{\(K_{6}(1)\)}}

\put(45,17){\makebox(0,0)[bl]{\(J\)}}
\put(45,13){\makebox(0,0)[bl]{\(J_{2}(2)\)}}
\put(45,09){\makebox(0,0)[bl]{\(J_{6}(1)\)}}

\put(30,17){\makebox(0,0)[bl]{\(G\)}}
\put(30,13){\makebox(0,0)[bl]{\(G_{2}(2)\)}}
\put(30,09){\makebox(0,0)[bl]{\(G_{3}(1)\)}}
\put(30,05){\makebox(0,0)[bl]{\(G_{6}(2)\)}}

\put(15,17){\makebox(0,0)[bl]{\(F\)}}
\put(15,13){\makebox(0,0)[bl]{\(F_{2}(2)\)}}
\put(15,09){\makebox(0,0)[bl]{\(F_{3}(1)\)}}

\put(00,17){\makebox(0,0)[bl]{\(E\)}}
\put(00,13){\makebox(0,0)[bl]{\(E_{2}(2)\)}}
\put(00,09){\makebox(0,0)[bl]{\(E_{3}(1)\)}}
\put(00,05){\makebox(0,0)[bl]{\(E_{6}(2)\)}}

\end{picture}
\end{center}

\newpage
\begin{center}
\begin{small}
 {\bf Table 7.} Compatibility factors for DAs pair (part \(B\))\\
\begin{tabular}{|c|l|l |}
\hline
  &DA \& DA & Factors \\

\hline

 1.&PK 1-4 E \& PK 9 F &Geological reserves, proximity\\

 2.&PK 1-4 E \& PK 10 G&Geological reserves, proximity \\

 3.&PK 1-4 E\& PK 11 J &Geological reserves, proximity\\

 4.&PK 9 F \& PK 10 G &Geological reserves, proximity\\

 5.&PK 9 F \& PK 11 J&Geological reserves, proximity\\

 6.&PK 10 G \& PK 11 J&Geological reserves, proximity\\

 7.&PK 12 K \& TP 1-2 L &Geological reserves, proximity\\

 8.&PK 12 K \& TP 5-5A V &Geological reserves, proximity\\

 9.&PK 12 K \& TP 10 O &Geological reserves, proximity\\

 10.&PK 12 K \& TP 11 P &Proximity, 'C5+v'\\

 11.& TP 1-2 L \& TP 5-5A V& Geological reserves, proximity, 'C5+v' \\

 12.& TP 1-2 L \& TP 10 O &Geological reserves, proximity, 'C5+v' \\

 13.& TP 1-2 L \& TP 11 P &Geological reserves, proximity, 'C5+v' \\

 14.&  TP 5-5A V \& TP 10 O &Geological reserves, proximity, 'C5+v' \\

 15.& TP 5-5A V \& TP 11 P &Geological reserves, proximity, 'C5+v' \\

 16.& TP 10 O \& TP 11 P  &Geological reserves, proximity, 'C5+v' \\

\hline
\end{tabular}
\end{small}
\end{center}

\begin{center}
\begin{small}
 {\bf Table 8.}  Compatibility for DAs (groups PK 1 - PK 11, part  \(B\) )\\
\begin{tabular}{|c|cc ccc ccc |}
\hline
 &\(F_{2}\)&\(F_{3}\)&\(G_{2}\)&\(G_{3}\)&\(G_{6}\)&\(J_{2}\)&\(J_{3}\)&\(J_{6}\)\\

\hline

 \(E_{2}\)&
 \(2\)&\(1\)&\(2\)&\(1\)&\(2\)&\(2\)&\(1\)&\(2\)\\

 \(E_{3}\)&
 \(4\)&\(3\)&\(4\)&\(3\)&\(1\)&\(4\)&\(3\)&\(1\)\\

 \(E_{6}\)&
 \(1\)&\(4\)&\(1\)&\(4\)&\(2\)&\(1\)&\(3\)&\(4\)\\

 \(F_{2}\)&
 \(\)&\(\)&\(3\)&\(4\)&\(2\)&\(3\)&\(4\)&\(2\)\\

 \(F_{3}\)&
 \(\)&\(\)&\(3\)&\(4\)&\(4\)&\(3\)&\(4\)&\(4\)\\

 \(G_{2}\)&
 \(\)&\(\)&\(\)&\(\)&\(\)&\(3\)&\(4\)&\(4\)\\

 \(G_{3}\)&
 \(\)&\(\)&\(\)&\(\)&\(\)&\(4\)&\(4\)&\(4\)\\

 \(G_{6}\)&
 \(\)&\(\)&\(\)&\(\)&\(\)&\(3\)&\(4\)&\(4\)\\

\hline
\end{tabular}
\end{small}
\end{center}

\begin{center}
\begin{small}
 {\bf Table 9.}  Compatibility for DAs (groups PK 12 - TP 11, part \(H\))\\
\begin{tabular}{|c|cc ccc ccc |}
\hline
 &\(L_{2}\)&\(L_{6}\)&\(V_{2}\)
 &\(V_{5}\)&\(O_{2}\)&\(O_{3}\)&\(P_{2}\)&\(P_{6}\)
 \\

\hline

 \(K_{2}\)&
 \(4\)&\(3\)&\(2\)&\(4\)&\(3\)&\(1\)&\(4\)&\(3\)\\

 \(K_{6}\)&
 \(1\)&\(4\)&\(3\)&\(4\)&\(3\)&\(4\)&\(3\)&\(4\)\\

   \(L_{2}\)&
 \(\)&\(\)&\(2\)&\(3\)&\(4\)&\(2\)&\(3\)&\(4\)\\

  \(L_{6}\)&
  \(\)&\(\)&\(2\)&\(4\)&\(4\)&\(4\)&\(3\)&\(4\)\\

 \(V_{2}\)&
 \(\)&\(\)&\(\)&\(\)&\(4\)&\(4\)&\(2\)&\(3\)\\

 \(V_{5}\)&
 \(\)&\(\)&\(\)&\(\)&\(3\)&\(4\)&\(3\)&\(4\)\\

 \(O_{2}\)&
 \(\)&\(\)&\(\)&\(\)&\(\)&\(\)&\(4\)&\(4\)\\

 \(O_{3}\)&
 \(\)&\(\)&\(\)&\(\)&\(\)&\(\)&\(3\)&\(4\)\\

\hline
\end{tabular}
\end{small}
\end{center}

\begin{center}
\begin{small}
 {\bf Table  10.} Intermediate composite DAs \\
\begin{tabular}{|l|c| }
\hline
  Composite DAs&\(N\)\\

\hline
 \(B_{1}=E_{3}\star F_{3}\star G_{3}\star J_{3}\)
 &\(3;4,0,0\)\\

 \(H_{1}=K_{6}\star L_{6}\star V_{5}\star O_{3}\star P_{6}\)
  &\(4;4,1,0\)\\

 \(H_{2}=K_{6}\star L_{6}\star V_{5}\star O_{3}\star P_{2}\)
  &\(3;5,0,0\)\\

\hline
\end{tabular}
\end{small}
\end{center}

\begin{center}
\begin{picture}(85,69)
\put(20,00){\makebox(0,0)[bl] {Fig. 10.
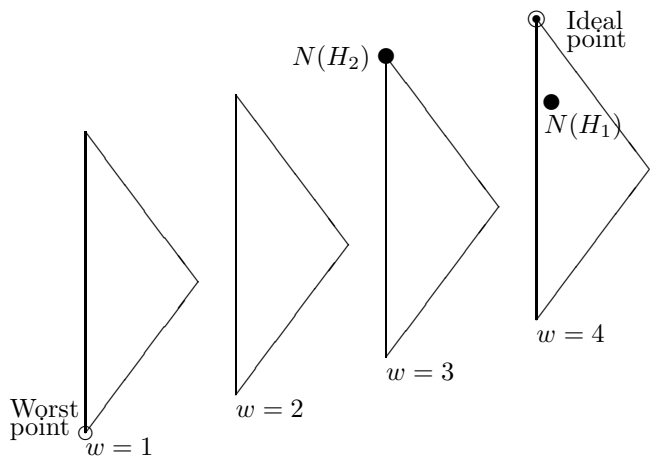
  Quality poset  for \(N(H)\)}}

\put(10,010){\line(0,1){40}} \put(10,010){\line(3,4){15}}
\put(10,050){\line(3,-4){15}}

\put(30,015){\line(0,1){40}} \put(30,015){\line(3,4){15}}
\put(30,055){\line(3,-4){15}}

\put(50,020){\line(0,1){40}} \put(50,020){\line(3,4){15}}
\put(50,060){\line(3,-4){15}}

\put(70,025){\line(0,1){40}} \put(70,025){\line(3,4){15}}
\put(70,065){\line(3,-4){15}}

\put(50,60){\circle*{2}}
\put(37.5,58){\makebox(0,0)[bl]{\(N(H_{2})\)}}


\put(72,54){\circle*{2}}
\put(71,49){\makebox(0,0)[bl]{\(N(H_{1})\)}}

\put(70,65){\circle{2}} \put(70,65){\circle*{1}}


\put(74,63.6){\makebox(0,0)[bl]{Ideal}}
\put(74,60.6){\makebox(0,0)[bl]{point}}

\put(10,07){\makebox(0,0)[bl]{\(w=1\)}}
\put(30,12){\makebox(0,0)[bl]{\(w=2\)}}
\put(50,17){\makebox(0,0)[bl]{\(w=3\)}}
\put(70,22){\makebox(0,0)[bl]{\(w=4\)}}

\put(10,10){\circle{1.7}}

\put(00,12){\makebox(0,0)[bl]{Worst}}
\put(00,09){\makebox(0,0)[bl]{point}}

\end{picture}
\end{center}

\subsection{Exploration plan for region}

  Thus, the following composite strategy for region
  is obtained (Fig. 11):


   {\bf 0.} General composite strategy
    \( S = A^{1} \star A^{2} \star A^{3} \star A^{4} \star A^{5}\)

   ~~{\bf 1.} Strategy for oil-gas field Kharosovey:
  \(A^{1}_{1}\).

  ~~{\bf 2.} Strategy for oil-gas field Arkticheskoe:
  \(A^{2}_{1}\), \(A^{2}_{2}\), \(A^{2}_{3}\), \(A^{2}_{4}\),
 \(A^{2}_{5}\), \(A^{2}_{6}\).

  ~~{\bf 3.} Strategy for oil-gas field Neitinskoe:
  \(A^{3}_{1}\).

  ~~{\bf 4.} Strategy for oil-gas field Kruzensternskoe:
  \(A^{4}_{1}\), \(A^{4}_{2}\).

  ~~{\bf 5.} Strategy for oil-gas field Bovanenkovskoe:
  \(A^{5}_{1}\), \(A^{5}_{2}\).



\begin{center}
\begin{picture}(75,62)
\put(01.5,00){\makebox(0,0)[bl]{Fig. 11.
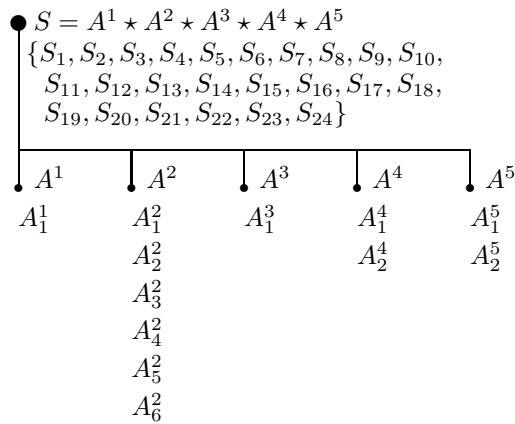
 Composite strategy for region }}

\put(00,58){\circle*{2}}

\put(00,41){\line(0,1){17}}

\put(02,57){\makebox(0,0)[bl] {\(S = A^{1} \star A^{2} \star A^{3}
\star A^{4} \star A^{5}\)}}

\put(01,52){\makebox(0,0)[bl]{\(\{S_{1},S_{2},S_{3},S_{4},S_{5},S_{6},S_{7},S_{8},S_{9},S_{10},
\)}}

\put(01,48){\makebox(0,0)[bl]{\(~~S_{11},S_{12},S_{13},S_{14},S_{15},S_{16},S_{17},S_{18},
\)}}

\put(01,44){\makebox(0,0)[bl]{\(~~S_{19},S_{20},S_{21},S_{22},S_{23},S_{24}\}
\)}}

\put(00,41){\line(1,0){60}}

\put(00,36){\line(0,1){05}} \put(15,36){\line(0,1){05}}
\put(30,36){\line(0,1){05}} \put(45,36){\line(0,1){05}}
\put(60,36){\line(0,1){05}}

\put(00,36){\circle*{1}} \put(15,36){\circle*{1}}
\put(30,36){\circle*{1}} \put(45,36){\circle*{1}}
\put(60,36){\circle*{1}}

\put(02,36){\makebox(0,0)[bl]{\(A^{1}\) }}
\put(17,36){\makebox(0,0)[bl]{\(A^{2}\)}}
\put(32,36){\makebox(0,0)[bl]{\(A^{3}\)}}
\put(47,36){\makebox(0,0)[bl]{\(A^{4}\)}}
\put(62,36){\makebox(0,0)[bl]{\(A^{5}\)}}

\put(60,30){\makebox(0,0)[bl]{\(A^{5}_{1}\)}}
\put(60,25){\makebox(0,0)[bl]{\(A^{5}_{2}\)}}

\put(45,30){\makebox(0,0)[bl]{\(A^{4}_{1}\)}}
\put(45,25){\makebox(0,0)[bl]{\(A^{4}_{2}\)}}

\put(30,30){\makebox(0,0)[bl]{\(A^{3}_{1}\)}}

\put(15,30){\makebox(0,0)[bl]{\(A^{2}_{1}\)}}
\put(15,25){\makebox(0,0)[bl]{\(A^{2}_{2}\)}}
\put(15,20){\makebox(0,0)[bl]{\(A^{2}_{3}\)}}
\put(15,15){\makebox(0,0)[bl]{\(A^{2}_{4}\)}}
\put(15,10){\makebox(0,0)[bl]{\(A^{2}_{5}\)}}
\put(15,05){\makebox(0,0)[bl]{\(A^{2}_{6}\)}}

\put(00,30){\makebox(0,0)[bl]{\(A^{1}_{1}\)}}

\end{picture}
\end{center}

 Finally, 24 composite exploration strategies for the
  region are
  (without compatibility analysis):

  \( S_{1} = A^{1}_{1} \star A^{2}_{1} \star A^{3}_{1} \star A^{4}_{1} \star
  A^{5}_{1}\),
  \( S_{2} = A^{1}_{1} \star A^{2}_{1} \star A^{3}_{1} \star A^{4}_{2} \star
  A^{5}_{1}\),

  \( S_{3} = A^{1}_{1} \star A^{2}_{1} \star A^{3}_{2} \star A^{4}_{1} \star
  A^{5}_{1}\),
  \( S_{4} = A^{1}_{1} \star A^{2}_{1} \star A^{3}_{2} \star A^{4}_{2} \star
  A^{5}_{1}\),

  \( S_{5} = A^{1}_{1} \star A^{2}_{2} \star A^{3}_{1} \star A^{4}_{1} \star
  A^{5}_{1}\),
  \( S_{6} = A^{1}_{1} \star A^{2}_{2} \star A^{3}_{1} \star A^{4}_{2} \star
  A^{5}_{1}\),

  \( S_{7} = A^{1}_{1} \star A^{2}_{2} \star A^{3}_{2} \star A^{4}_{1} \star
  A^{5}_{1}\),
  \( S_{8} = A^{1}_{1} \star A^{2}_{2} \star A^{3}_{2} \star A^{4}_{2} \star
  A^{5}_{1}\),

   \( S_{9} = A^{1}_{1} \star A^{2}_{3} \star A^{3}_{1} \star A^{4}_{1} \star
  A^{5}_{1}\),
  \( S_{10} = A^{1}_{1} \star A^{2}_{3} \star A^{3}_{1} \star A^{4}_{2} \star
  A^{5}_{1}\),

  \( S_{11} = A^{1}_{1} \star A^{2}_{3} \star A^{3}_{2} \star A^{4}_{1} \star
  A^{5}_{1}\),
  \( S_{12} = A^{1}_{1} \star A^{2}_{3} \star A^{3}_{2} \star A^{4}_{2} \star
  A^{5}_{1}\),

    \( S_{13} = A^{1}_{1} \star A^{2}_{4} \star A^{3}_{1} \star A^{4}_{1} \star
  A^{5}_{1}\),
  \( S_{14} = A^{1}_{1} \star A^{2}_{4} \star A^{3}_{1} \star A^{4}_{2} \star
  A^{5}_{1}\),

  \( S_{15} = A^{1}_{1} \star A^{2}_{4} \star A^{3}_{2} \star A^{4}_{1} \star
  A^{5}_{1}\),
  \( S_{16} = A^{1}_{1} \star A^{2}_{4} \star A^{3}_{2} \star A^{4}_{2} \star
  A^{5}_{1}\),

   \( S_{17} = A^{1}_{1} \star A^{2}_{5} \star A^{3}_{1} \star A^{4}_{1} \star
  A^{5}_{1}\),
  \( S_{18} = A^{1}_{1} \star A^{2}_{5} \star A^{3}_{1} \star A^{4}_{2} \star
  A^{5}_{1}\),

  \( S_{19} = A^{1}_{1} \star A^{2}_{5} \star A^{3}_{2} \star A^{4}_{1} \star
  A^{5}_{1}\),
  \( S_{20} = A^{1}_{1} \star A^{2}_{5} \star A^{3}_{2} \star A^{4}_{2} \star
  A^{5}_{1}\),

   \( S_{21} = A^{1}_{1} \star A^{2}_{6} \star A^{3}_{1} \star A^{4}_{1} \star
  A^{5}_{1}\),
  \( S_{22} = A^{1}_{1} \star A^{2}_{6} \star A^{3}_{1} \star A^{4}_{2} \star
  A^{5}_{1}\),

  \( S_{23} = A^{1}_{1} \star A^{2}_{6} \star A^{3}_{2} \star A^{4}_{1} \star
  A^{5}_{1}\),
  \( S_{24} = A^{1}_{1} \star A^{2}_{6} \star A^{3}_{2} \star A^{4}_{2} \star
  A^{5}_{1}\).

 Now an additional analysis of the obtained strategies can be
 considered to design the best final strategy
 (e.g., multicriteria analysis and selection, expert judgment).
 On the other hand, the final strategy can be build by
 aggregation of the obtained solutions.

\subsection{Aggregation of solutions}

  In the considered example,
  there are 24 solutions (previous section):
 ~\(S_{1}\),...,\(S_{24}\).
 The substructure of the solutions is shown in Fig. 12.
 This structure is used as a 'kernel'  for an extension process.
 The superstructure is shown in Fig. 13.

\begin{center}
\begin{picture}(70,20)
\put(0,00){\makebox(0,0)[bl] {Fig. 12. Substructure ('kernel')}}

\put(05,14){\circle*{1.5}} \put(13,14){\circle*{1.5}}
\put(21,14){\circle*{1.5}} \put(29,14){\circle*{1.5}}
\put(37,14){\circle*{1.5}}

\put(04,16){\makebox(0,0)[bl]{\(A^{1}\)}}
\put(12,16){\makebox(0,0)[bl]{\(A^{2}\)}}
\put(20,16){\makebox(0,0)[bl]{\(A^{3}\)}}
\put(28,15.5){\makebox(0,0)[bl]{\(A^{4}\)}}

\put(35,16){\makebox(0,0)[bl]{\(A^{5}\)}}

\put(05,10){\oval(07,08)} \put(13,10){\oval(07,08)}
\put(21,10){\oval(07,08)} \put(29,10){\oval(07,08)}
\put(37,10){\oval(07,08)}

\put(02.8,08){\makebox(0,0)[bl]{\(A^{1}_{1}\)}}

\put(18.8,08){\makebox(0,0)[bl]{\(A^{3}_{1}\)}}

\end{picture}
%
\begin{picture}(42,42)

\put(02.5,00){\makebox(0,0)[bl] {Fig. 13. Superstructure}}

\put(05,36){\circle*{1.5}} \put(13,36){\circle*{1.5}}
\put(21,36){\circle*{1.5}} \put(29,36){\circle*{1.5}}
\put(37,36){\circle*{1.5}}

\put(04,38){\makebox(0,0)[bl]{\(A^{1}\)}}
\put(12,38){\makebox(0,0)[bl]{\(A^{2}\)}}
\put(20,38){\makebox(0,0)[bl]{\(A^{3}\)}}
\put(28,38){\makebox(0,0)[bl]{\(A^{4}\)}}
\put(35,38){\makebox(0,0)[bl]{\(A^{5}\)}}


\put(05,20.5){\oval(07,31)} \put(13,20.5){\oval(07,31)}
\put(21,20.5){\oval(07,31)} \put(29,20.5){\oval(07,31)}
\put(37,20.5){\oval(07,31)}

\put(02.8,31){\makebox(0,0)[bl]{\(A^{1}_{1}\)}}

\put(10.8,31){\makebox(0,0)[bl]{\(A^{2}_{1}\)}}
\put(10.8,26){\makebox(0,0)[bl]{\(A^{2}_{2}\)}}
\put(10.8,21){\makebox(0,0)[bl]{\(A^{2}_{3}\)}}
\put(10.8,16){\makebox(0,0)[bl]{\(A^{2}_{4}\)}}
\put(10.8,11){\makebox(0,0)[bl]{\(A^{2}_{5}\)}}
\put(10.8,06){\makebox(0,0)[bl]{\(A^{2}_{6}\)}}

\put(18.8,31){\makebox(0,0)[bl]{\(A^{3}_{2}\)}}

\put(26.8,31){\makebox(0,0)[bl]{\(A^{4}_{1}\)}}
\put(26.8,26){\makebox(0,0)[bl]{\(A^{4}_{2}\)}}

\put(34.8,31){\makebox(0,0)[bl]{\(A^{5}_{1}\)}}
\put(34.8,26){\makebox(0,0)[bl]{\(A^{5}_{2}\)}}

\end{picture}
\end{center}

 Table 11 contains design alternatives for extension of the
 'kernel' including their estimates
 (ordinal scales are used, expert judgment).

\begin{center}
\begin{small}
 {\bf Table  11.} Extension versions \\
\begin{tabular}{|c|c|c|c|c| }
\hline
 \(\kappa\)& Versions&Binary&Cost      &Profit\\
           &      & variable&\(a_{ij}\)&\(c_{ij}\)\\

\hline

  1.&\(A^{2}_{1}\)&\(x_{11}\)&\(4\)&\(4\)\\
  2.&\(A^{2}_{2}\)&\(x_{12}\)&\(6\)&\(6\)\\
  3.&\(A^{2}_{3}\)&\(x_{13}\)&\(3\)&\(2\)\\
  4.&\(A^{2}_{4}\)&\(x_{14}\)&\(3\)&\(3\)\\
  5.&\(A^{2}_{5}\)&\(x_{15}\)&\(4\)&\(3\)\\
  6.&\(A^{2}_{6}\)&\(x_{16}\)&\(5\)&\(3\)\\
  7.&\(A^{4}_{1}\)&\(x_{21}\)&\(3\)&\(4\)\\
  8.&\(A^{4}_{2}\)&\(x_{22}\)&\(3\)&\(3\)\\
  9.&\(A^{5}_{1}\)&\(x_{31}\)&\(3\)&\(3\)\\
 10.&\(A^{5}_{2}\)&\(x_{32}\)&\(4\)&\(4\)\\

\hline
\end{tabular}
\end{small}
\end{center}

 It is assumed, the DAs are compatible.
 The aggregation problem (extension strategy)
 is based on multiple choice problem:
 \[\max \sum_{i=1}^{3}  \sum_{j=1}^{q_{i}}   c_{ij} x_{ij}
%
 ~~~~~~ s.t.~ \sum_{i=1}^{3}  \sum_{j=1}^{q_{i}}   a_{ij} x_{ij} \leq
 b,
 ~~\sum_{j=1}^{q_{i}}   x_{ij} = 1 ~~  \forall i=\overline{1,3},
  ~~x_{ij} \in \{0,1\}.\]
%
%
 In this model,
 \(q_{1} = 6\), \(q_{2} = 2\), \(q_{3} = 2\).
 By the usage of a greedy algorithm
 (i.e., linear ordering of elements by
  \(c_{i}/a_{i}\))
  the following solutions are obtained
  for four versions of constraints:

  (1) \(b^{1}=9\):~
 (\(x_{14} = 10\), \(x_{21} = 1\), \(x_{31} = 1\)),

 ~\(S^{agg}_{b^{1}} = A^{1}_{1} \star A^{2}_{4} \star A^{3}_{1}
 \star A^{4}_{1} \star A^{5}_{1} =
 R_{3}\star P_{3} \star D_{2}\star Q_{4}\star U_{1}\star Z_{1}
 \star Y_{2}\star O_{1} \);

 (2.1.) \(b^{2}=10\):~
 (\(x_{14} = 1\), \(x_{21} = 1\), \(x_{32} = 1\)),

 ~\(S^{agg1}_{b^{2}} =
 A^{1}_{1} \star A^{2}_{4} \star A^{3}_{1}
 \star A^{4}_{1} \star A^{5}_{2} =
 R_{3}\star P_{3} \star D_{2}\star Q_{4}\star U_{1}\star Z_{1}
 \star Y_{2}\star O_{1} \);

 (2.2.) \(b^{2}=10\):~
 (\(x_{11} = 1\), \(x_{21} = 1\), \(x_{31} = 1\)),

 ~\(S^{agg2}_{b^{2}} =
 A^{1}_{1} \star A^{2}_{1} \star A^{3}_{1}
 \star A^{4}_{1} \star A^{5}_{1} =
 R_{3}\star P_{3} \star D_{2}\star Q_{4}\star U_{1}\star Z_{1}
 \star Y_{2}\star O_{1} \);

 (3) \(b^{3}=11\):~
 (\(x_{11} = 1\), \(x_{21} = 1\), \(x_{32} = 1\)),

 ~\(S^{agg}_{b^{3}} =
 A^{1}_{1} \star A^{2}_{1} \star A^{3}_{1}
 \star A^{4}_{1} \star A^{5}_{2} =
 R_{4}\star P_{3} \star D_{2}\star Q_{4}\star U_{1}\star Z_{1}
 \star Y_{2}\star O_{1} \);

(4) \(b^{4}=11\):~
 (\(x_{12} = 1\), \(x_{22} = 1\), \(x_{31} = 1\)),

 ~\(S^{agg}_{b^{4}} =
 A^{1}_{1} \star A^{2}_{2} \star A^{3}_{1}
 \star A^{4}_{2} \star A^{5}_{1} =
 R_{4}\star P_{3} \star D_{2}\star Q_{4}\star U_{1}\star Z_{1}
 \star Y_{2}\star O_{1} \).

\subsection{Example of multiset estimates based synthesis}

  A scale based on multiset estimates
  (as a poset) for the used assessment problem
 \(P^{3,4}\) is depicted in Fig. 14.
 The illustrative  numerical example is based on multiset
 estimates for Arkticheskoe oil-gas field (Fig. 15).
 Multiset estimates for local DAs are shown in Fig. 15
 (in parentheses).
 Compatibility estimates from Table 3 are used.
 Two solutions are considered:

 \(W^{M}_{1} = E_{6}\star F_{6}\star G_{6}\star J_{6}\star I_{6}\),
 ~ \(N(W^{M}_{1}) = (w(W^{M}_{1});e(W^{M}_{1})) = (4;1,3,0)\);

 \(W^{M}_{2} = E_{6}\star F_{6}\star G_{3}\star J_{6}\star I_{3}\),
 ~ \(N(W^{M}_{2}) = (w(W^{M}_{2});e(W^{M}_{2})) = (3;3,1,0)\).

 Estimates
 \( e(W^{M}_{1}) = (1,3,0)\),
 \( e(W^{M}_{2}) = (3,1,0)\)
 are medians for estimates of the corresponding components.

\begin{center}
\begin{picture}(78,114)
\put(00,00){\makebox(0,0)[bl] {Fig. 14.
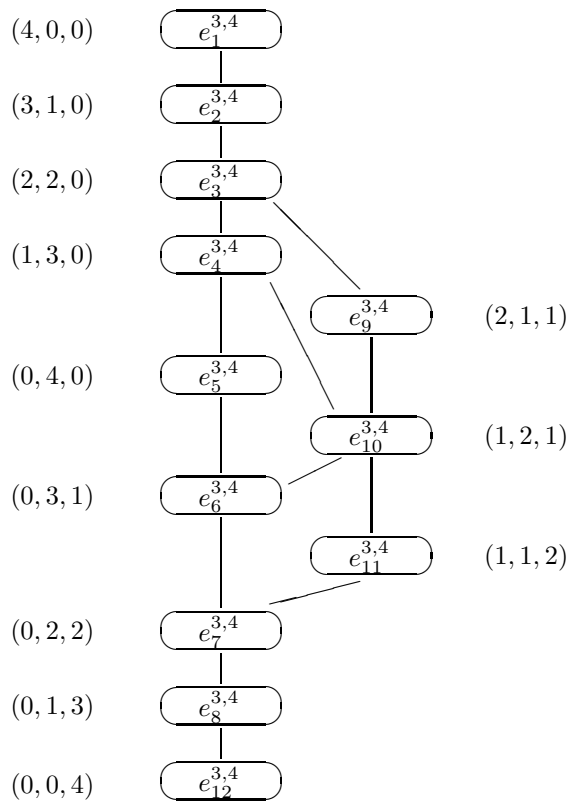
 Estimates for assessment problem  \(P^{3,4}\)}}

\put(25,106.7){\makebox(0,0)[bl]{\(e^{3,4}_{1}\) }}

\put(28,109){\oval(16,5)}


\put(00,107){\makebox(0,0)[bl]{\((4,0,0)\) }}

\put(28,102){\line(0,1){4}}

\put(25,96.7){\makebox(0,0)[bl]{\(e^{3,4}_{2}\) }}

\put(28,99){\oval(16,5)}

\put(00,97){\makebox(0,0)[bl]{\((3,1,0)\) }}

\put(28,92){\line(0,1){4}}

\put(25,86.7){\makebox(0,0)[bl]{\(e^{3,4}_{3}\) }}

\put(28,89){\oval(16,5)}

\put(00,87){\makebox(0,0)[bl]{\((2,2,0)\) }}


\put(28,82){\line(0,1){4}}

\put(25,76.7){\makebox(0,0)[bl]{\(e^{3,4}_{4}\) }}

\put(28,79){\oval(16,5)}

\put(00,77){\makebox(0,0)[bl]{\((1,3,0)\) }}

\put(28,66){\line(0,1){10}}

\put(25,60.7){\makebox(0,0)[bl]{\(e^{3,4}_{5}\) }}

\put(28,63){\oval(16,5)}

\put(00,61){\makebox(0,0)[bl]{\((0,4,0)\) }}

\put(28,50){\line(0,1){10}}

\put(25,44.7){\makebox(0,0)[bl]{\(e^{3,4}_{6}\) }}

\put(28,47){\oval(16,5)}

\put(00,45){\makebox(0,0)[bl]{\((0,3,1)\) }}

\put(28,32){\line(0,1){12}}

\put(25,26.7){\makebox(0,0)[bl]{\(e^{3,4}_{7}\) }}

\put(28,29){\oval(16,5)}

\put(00,27){\makebox(0,0)[bl]{\((0,2,2)\) }}

\put(28,22){\line(0,1){4}}

\put(25,16.7){\makebox(0,0)[bl]{\(e^{3,4}_{8}\) }}

\put(28,19){\oval(16,5)}

\put(00,17){\makebox(0,0)[bl]{\((0,1,3)\) }}

\put(28,12){\line(0,1){4}}

\put(25,06.7){\makebox(0,0)[bl]{\(e^{3,4}_{12}\) }}

\put(28,9){\oval(16,5)}

\put(00,06.5){\makebox(0,0)[bl]{\((0,0,4)\) }}

\put(46.5,74.5){\line(-1,1){11.5}}

\put(45,68.7){\makebox(0,0)[bl]{\(e^{3,4}_{9}\) }}

\put(48,71){\oval(16,5)}

\put(63,69){\makebox(0,0)[bl]{\((2,1,1)\) }}

\put(43,58.5){\line(-1,2){8.4}}

\put(44,52){\line(-2,-1){7}}

\put(48,58){\line(0,1){10}}

\put(45,52.7){\makebox(0,0)[bl]{\(e^{3,4}_{10}\) }}

\put(48,55){\oval(16,5)}

\put(63,53){\makebox(0,0)[bl]{\((1,2,1)\) }}

\put(46.5,35.5){\line(-4,-1){12}}

\put(48,42){\line(0,1){10}}

\put(45,36.7){\makebox(0,0)[bl]{\(e^{3,4}_{11}\) }}

\put(48,39){\oval(16,5)}

\put(63,37){\makebox(0,0)[bl]{\((1,1,2)\) }}

\end{picture}
\end{center}

\begin{center}
\begin{picture}(96,46)
\put(03.6,00){\makebox(0,0)[bl]{Fig. 15.
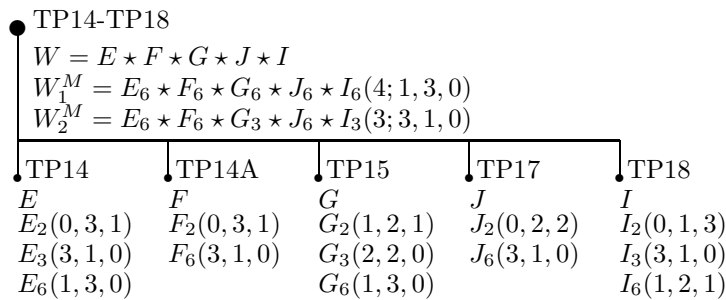
 Arkticheskoe oil-gas field (multiset esitmates)}}

\put(02,41){\makebox(0,0)[bl]{TP14-TP18}}

\put(00,41){\circle*{2}}

\put(00,22){\line(0,1){19}}

\put(02,36){\makebox(0,0)[bl]{\(W=E\star F\star G\star J \star I
\)}}


\put(02,31){\makebox(0,0)[bl]{\(W^{M}_{1}=E_{6}\star F_{6}\star
G_{6} \star J_{6}\star I_{6} (4;1,3,0)\)}}


\put(02,27){\makebox(0,0)[bl]{\(W^{M}_{2}=E_{6}\star F_{6}\star
G_{3} \star J_{6}\star I_{3} (3;3,1,0)\)}}

\put(00,26){\line(1,0){80}}

\put(00,21){\line(0,1){05}} \put(20,21){\line(0,1){05}}
\put(40,21){\line(0,1){05}} \put(60,21){\line(0,1){05}}
\put(80,21){\line(0,1){05}}

\put(00,21){\circle*{1}} \put(20,21){\circle*{1}}
\put(40,21){\circle*{1}} \put(60,21){\circle*{1}}
\put(80,21){\circle*{1}}

\put(01,21){\makebox(0,0)[bl]{TP14}}

\put(21,21){\makebox(0,0)[bl]{TP14A}}

\put(41,21){\makebox(0,0)[bl]{TP15}}

\put(61,21){\makebox(0,0)[bl]{TP17}}

\put(81,21){\makebox(0,0)[bl]{TP18}}


\put(80,17){\makebox(0,0)[bl]{\(I\)}}
\put(80,13){\makebox(0,0)[bl]{\(I_{2}(0,1,3)\)}}
\put(80,09){\makebox(0,0)[bl]{\(I_{3}(3,1,0)\)}}
\put(80,05){\makebox(0,0)[bl]{\(I_{6}(1,2,1)\)}}

\put(60,17){\makebox(0,0)[bl]{\(J\)}}
\put(60,13){\makebox(0,0)[bl]{\(J_{2}(0,2,2)\)}}
\put(60,09){\makebox(0,0)[bl]{\(J_{6}(3,1,0)\)}}

\put(40,17){\makebox(0,0)[bl]{\(G\)}}
\put(40,13){\makebox(0,0)[bl]{\(G_{2}(1,2,1)\)}}
\put(40,09){\makebox(0,0)[bl]{\(G_{3}(2,2,0)\)}}
\put(40,05){\makebox(0,0)[bl]{\(G_{6}(1,3,0)\)}}

\put(20,17){\makebox(0,0)[bl]{\(F\)}}
\put(20,13){\makebox(0,0)[bl]{\(F_{2}(0,3,1)\)}}
\put(20,09){\makebox(0,0)[bl]{\(F_{6}(3,1,0)\)}}

\put(00,17){\makebox(0,0)[bl]{\(E\)}}
\put(00,13){\makebox(0,0)[bl]{\(E_{2}(0,3,1)\)}}
\put(00,09){\makebox(0,0)[bl]{\(E_{3}(3,1,0)\)}}
\put(00,05){\makebox(0,0)[bl]{\(E_{6}(1,3,0)\)}}

\end{picture}
\end{center}

\section{Conclusion}

 This paper describes a hierarchical approach
 to
 combinatorial planning
  of geological exploration.
%
 The approach is based on the following:
 (a) expert judgment;
%
 (b) planning consists in
 bottom-up selection and composition
 of local solutions (design/exploration alternatives DAs) into
 composite solutions at the higher layer of the plan hierarchy;
%
 (c) aggregation of the obtained plans (solutions)
  is considered as an extension of a 'kernel'
  of the preliminary obtained solution versions.
 The approach is illustrated by a numerical example
 as oil and gas geological planning
 for Yamal peninsula.
%
 It may be reasonable
 to consider the following future directions:
 (1) examination of multistage exploration strategies;
 (2) study of combinatorial evolution models for
  oil and gas field(s);
  (3) using the suggested framework
 in education.


\end{document}